\documentclass[colorlinks=false]{jmlrcia} 

\usepackage{microtype}
\usepackage{booktabs}

\usepackage{tikz}
    \definecolor{mycolor1}{HTML}{009900}
    \definecolor{mycolor2}{HTML}{990000}
    \usetikzlibrary{decorations.pathmorphing}
	\tikzstyle{cut-edge}=[dotted]
    \tikzstyle{vertex}=[circle, draw, fill=white, inner sep=0pt, minimum width=1ex]
    \tikzstyle{dot}=[circle, draw, fill=white, inner sep=0pt, minimum width=1ex]
    \tikzset{every picture/.append style={baseline,scale=1.1}}
\usepackage{pgfplots}

\newcommand{\conv}{\mathrm{conv}\,}
\newcommand{\sgn}{\mathrm{sgn}\,}

\def\clap#1{\hbox to 0pt{\hss#1\hss}}

\newcommand{\smallplotwidth}{0.33\textwidth}

\title[Convexification of Learning from Constraints]{Convexification of Learning from Constraints}
\usepackage{times}

 \coltauthor{\Name{Iaroslav Shcherbatyi} \Email{shcherbatyi@mpi-inf.mpg.de}\\
 \addr Max Planck Institute for Informatics, Saarbr\"ucken, Germany
 \AND
 \Name{Bjoern Andres} \Email{andres@mpi-inf.mpg.de}\\
 \addr Max Planck Institute for Informatics, Saarbr\"ucken, Germany\\[-5ex] 
 }

\begin{document}

\maketitle

\begin{abstract}
Regularized empirical risk minimization with constrained labels (in contrast to fixed labels) is a remarkably general abstraction of learning.
For common loss and regularization functions, this optimization problem assumes the form of a mixed integer program (MIP) whose objective function is non-convex. 
In this form, the problem is resistant to standard optimization techniques.
We construct MIPs with the same solutions whose objective functions are convex.
Specifically, we characterize the tightest convex extension of the objective function, 
given by the Legendre-Fenchel biconjugate.
Computing values of this tightest convex extension is NP-hard.
However, by applying our characterization to every function in an additive decomposition of the objective function, we obtain a class of looser convex extensions that can be computed efficiently.
For some decompositions, common loss and regularization functions, we derive a closed form.
\end{abstract}


\section{Introduction}

We study an optimization problem:
Given 
a finite set $S \neq \emptyset$ whose elements are to be labeled as 0 or 1,
a non-empty set $Y \subseteq \{0,1\}^S$ of \emph{feasible labelings},
$x: S \to \mathbb{R}^m$ with $m \in \mathbb{N}$, called a \emph{feature matrix},
$l: \mathbb{R} \times [0,1] \to \mathbb{R}_0^+$ with 
$l(\cdot, 0)$ convex,
$l(\cdot, 1)$ convex,
$l(r, 1) \rightarrow 0$ as $r \rightarrow \infty$ 
and $l(-r, 0) \rightarrow 0$ as $r \rightarrow \infty$, 
called a \emph{loss function},
$\Theta \subseteq \mathbb{R}^m$ convex, called the set of \emph{feasible parameters},
$\omega: \Theta \to \mathbb{R}_0^+$ convex, called a \emph{regularization function},
and $C \in \mathbb{R}_0^+$, called a \emph{regularization constant},
we consider the optimization problem 
\begin{align}
\inf_{(\theta, y) \in \Theta \times Y} 
	\varphi(\theta, y)
\label{eq:problem-main}
\end{align}
\vspace{-1ex}
with
\vspace{-1ex}
\begin{align}
\varphi(\theta, y)\ :=\ 
    & \omega(\theta) + \frac{C}{|S|} \sum_{s \in S} l_s(\theta, y_s) \\
l_s(\theta, y_s)\ :=\ 
    & l(\langle x_s, \theta \rangle, y_s) 
\enspace .
\end{align}
A minimizer $(\hat\theta, \hat y)$, if it exists, defines a \emph{classifier}
$c: \mathbb{R}^m \to \{0,1\}: x \mapsto \frac{1}{2}(1 + \sgn \langle \hat\theta, x \rangle)$ 
and a feasible labeling $\hat y \in Y$.
The optimization problem \eqref{eq:problem-main} is a remarkably general abstraction of learning.
On the one hand, it generalizes supervised, semi-supervised and unsupervised learning:
\begin{itemize}
\itemsep0ex
\item If the labeling $y$ is fixed by $|Y| = 1$ to precisely one feasible labeling,
\eqref{eq:problem-main} 
specializes to \emph{regularized empirical risk minimization with fixed labels} and is called \emph{supervised}.
\item If $Y$ fixes the label of at least one but not all elements of $S$, \eqref{eq:problem-main} is called \emph{semi-supervised}.
\item If $Y$ constrains the joint labelings of $S$ without fixing the label of any single element of $S$, \eqref{eq:problem-main} is called \emph{unsupervised}.
For example, consider $Y = \{y \in \{0,1\}^S\ |\ \sum_{s \in S} y_s = \lfloor |S|/2 \rfloor \}$.
\end{itemize}
On the other hand, the optimization problem \eqref{eq:problem-main} generalizes classification, clustering and ranking:
\begin{itemize}
\itemsep0ex
\item If $S = A \times B$ and $Y$ is the set of characteristic functions of \emph{maps} from $A$ to $B$, \eqref{eq:problem-main} is an abstraction of \emph{multi-label classification}, as discussed, for instance, by 
\citet{joachims-1999,joachims-2003,chapelle-2005-semi,xu-2005-multi-class,chapelle-2006-branch,chapelle-2006-continuation,chapelle-2008-optimization}.
\item If $S = A \times A$ and $Y$ is the set of characteristic functions of \emph{equivalence relations} on $A$, \eqref{eq:problem-main} is an abstraction of \emph{clustering}, as discussed by 
\cite{finley-2005,xu-2005-clustering}. 
For fixed parameters $\theta$, 
it specializes to the NP-hard minimum cost clique partition problem
\citep{groetschel-1989,chopra-1993}
that is also known as correlation clustering
\citep{bansal2004correlation,demaine-2006}.
\item If $S = A \times A$ and $Y$ is the set of characteristic functions of \emph{linear orders} on $A$, \eqref{eq:problem-main} is an abstraction of \emph{ranking}. 
For fixed parameters $\theta$, 
it specializes to the NP-hard linear ordering problem
\citep{marti-2011}.
\end{itemize}

The set $Y$ of feasible labelings, a subset of the vertices of the unit hypercube $[0,1]^S$, is a non-convex subset of $\mathbb{R}^S$ iff $2 \leq |Y|$.
Typically, one considers a relaxation $y \in P$ of the constraint $y \in Y$ with $P$ a convex polytope such that $\conv\,Y \subseteq P \subseteq [0,1]^S$ and $P \cap \{0,1\}^S = Y$.
In practice, one considers a polytope that is described as an intersection of half-spaces, i.e., by $n \in \mathbb{N}$, $A \in \mathbb{R}^{n \times S}$ and $b \in \mathbb{R}^n$ such that $P = \{y \in \mathbb{R}^S\,|\,A y \leq b \}$.
Also typically, the set $\Theta \subseteq \mathbb{R}^m$ of feasible parameters is a convex polyhedron and is described also as an intersection of half-spaces, i.e., by $n' \in \mathbb{N}$, $A' \subseteq \mathbb{R}^{n' \times m}$ and $b' \in \mathbb{R}^{n'}$ such that $\Theta = \{\theta \in \mathbb{R}^m\,|\,A' \theta \leq b'\}$.
Hence, the optimization problem 
\eqref{eq:problem-main}
assumes the form of a mixed integer program (MIP):
\begin{align}
\inf_{(\theta, y) \in \mathbb{R}^m \times [0,1]^S} \quad
    & \varphi(\theta, y) 
    \label{eq:mip-objective}\\
\textnormal{subject to} \quad
    & A' \theta \leq b' \label{eq:mpi-theta-polytope} \\
    & A y \leq b \label{eq:mip-label-polytope}\\
    & y \in \{0,1\}^S \label{eq:mip-integrality}
\end{align}

\begin{table}[b]
\vspace{-2ex} 
\centering
\small
\caption{Loss functions}
\label{table:loss-functions}
\vspace{1ex}
\begin{tabular}{lll}
\toprule
Loss 
	& Form of $l_s(\theta, y_s)$ 
	& Function $l_s: \mathbb{R}^m \times [0,1] \to \mathbb{R}_0^+$ \\
\midrule
Squared difference 
	& $(\langle \theta, x_s \rangle - y_s)^2$ 
	& convex \\
Logistic 
	& $\log(1 + \exp(-(2 y_s - 1) \langle \theta, x_s \rangle))$ 
	& non-convex \\
Hinge 
	& $\max \{0, 1 - (2 y_s - 1) \langle \theta, x_s \rangle\}$ 
	& non-convex \\	
Squared Hinge 
	& $\max \{0, 1 - (2 y_s - 1) \langle \theta, x_s \rangle\}^2$
	& non-convex \\
\bottomrule
\end{tabular}
\end{table}
For \emph{convex} loss functions $l_s$ such as the squared difference loss 
(Tab.~\ref{table:loss-functions}),
the objective function $\varphi$ is convex on the domain $\mathbb{R}^m \times [0,1]^S$.
Thus, the continuous relaxation \eqref{eq:mip-objective}--\eqref{eq:mip-label-polytope}
of the problem \eqref{eq:mip-objective}--\eqref{eq:mip-integrality}
is a convex problem.
Its solutions $(\hat \theta', \hat y')$, although possibly fractional in the coordinates of $y'$, can inform a search for feasible solutions of 
\eqref{eq:mip-objective}--\eqref{eq:mip-integrality}
with certificates
\citep{chapelle-2006-branch,chapelle-2008-optimization,bojanowski2013finding}. 
See also
\citet{bonami2012algorithms}
for a recent survey of convex mixed-integer non-linear programming.
For \emph{non-convex} loss functions $l_s$ such as the logistic loss, the Hinge loss and the squared Hinge loss
(Tab.~\ref{table:loss-functions}),
$\varphi$ is non-convex on the domain $\mathbb{R}^m \times [0,1]^S$.
In this case, 
\eqref{eq:mip-objective}--\eqref{eq:mip-integrality}
is resistant to standard optimization techniques.
See
\citet{tawarmalani-2004,lee-2011,belotti-2013}
for an overview of non-convex mixed-integer non-linear programming.

\subsection{Contribution}
\begin{figure*}
\subfigure[]{
\label{fig:mip-non-convex}
%
%
\begin{tikzpicture}

\begin{axis}[%
    width=0.33\textwidth,
    xmin=-3,
    xmax=3,
    tick align=outside,
    xlabel={$\theta$},
    xlabel style={yshift=3ex},
    xmajorgrids,
    ymin=0,
    ymax=1,
    ylabel={y},
    ylabel style={yshift=3ex},
    ymajorgrids,
    zmin=0.72,
    zmax=24.5,
    zmajorgrids,
    view={-23.5}{30},
    axis x line*=bottom,
    axis y line*=left,
    axis z line*=left,
    colormap={mycolormap}{color(0)=(white) color(1)=(mycolor1)}
]

\addplot3[solid,line width=2.0pt]
 table[row sep=crcr] {%
-3	0	4.5\\
-2.7	0	3.645\\
-2.4	0	2.88\\
-2.1	0	2.205\\
-1.8	0	1.62\\
-1.5	0	1.125\\
-1.2	0	0.72\\
-0.9	0	0.905\\
-0.6	0	2.18\\
-0.3	0	3.545\\
0	0	5\\
0.3	0	6.545\\
0.6	0	8.18\\
0.9	0	9.905\\
1.2	0	11.72\\
1.5	0	13.625\\
1.8	0	15.62\\
2.1	0	17.705\\
2.4	0	19.88\\
2.7	0	22.145\\
3	0	24.5\\
};

\addplot3[solid,line width=2.0pt]
 table[row sep=crcr] {%
-3	1	24.5\\
-2.7	1	22.145\\
-2.4	1	19.88\\
-2.1	1	17.705\\
-1.8	1	15.62\\
-1.5	1	13.625\\
-1.2	1	11.72\\
-0.9	1	9.905\\
-0.6	1	8.18\\
-0.3	1	6.545\\
0	1	5\\
0.3	1	3.545\\
0.6	1	2.18\\
0.9	1	0.905\\
1.2	1	0.72\\
1.5	1	1.125\\
1.8	1	1.62\\
2.1	1	2.205\\
2.4	1	2.88\\
2.7	1	3.645\\
3	1	4.5\\
};

\addplot3[%
    surf,
    shader=faceted,
    mesh/rows=21
]
table[
    row sep=crcr,
    header=false
] {%
-3	0.005	4.5\\
-3	0.0545	4.5\\
-3	0.104	4.5\\
-3	0.1535	4.5\\
-3	0.203	4.5\\
-3	0.2525	4.5\\
-3	0.302	4.5\\
-3	0.3515	5.045\\
-3	0.401	6.53\\
-3	0.4505	8.015\\
-3	0.5	9.5\\
-3	0.5495	10.985\\
-3	0.599	12.47\\
-3	0.6485	13.955\\
-3	0.698	15.44\\
-3	0.7475	16.925\\
-3	0.797	18.41\\
-3	0.8465	19.895\\
-3	0.896	21.38\\
-3	0.9455	22.865\\
-3	0.995	24.35\\
-2.7	0.005	3.645\\
-2.7	0.0545	3.645\\
-2.7	0.104	3.645\\
-2.7	0.1535	3.645\\
-2.7	0.203	3.645\\
-2.7	0.2525	3.645\\
-2.7	0.302	3.645\\
-2.7	0.3515	4.6355\\
-2.7	0.401	5.972\\
-2.7	0.4505	7.3085\\
-2.7	0.5	8.645\\
-2.7	0.5495	9.9815\\
-2.7	0.599	11.318\\
-2.7	0.6485	12.6545\\
-2.7	0.698	13.991\\
-2.7	0.7475	15.3275\\
-2.7	0.797	16.664\\
-2.7	0.8465	18.0005\\
-2.7	0.896	19.337\\
-2.7	0.9455	20.6735\\
-2.7	0.995	22.01\\
-2.4	0.005	2.88\\
-2.4	0.0545	2.88\\
-2.4	0.104	2.88\\
-2.4	0.1535	2.88\\
-2.4	0.203	2.88\\
-2.4	0.2525	2.88\\
-2.4	0.302	3.128\\
-2.4	0.3515	4.316\\
-2.4	0.401	5.504\\
-2.4	0.4505	6.692\\
-2.4	0.5	7.88\\
-2.4	0.5495	9.068\\
-2.4	0.599	10.256\\
-2.4	0.6485	11.444\\
-2.4	0.698	12.632\\
-2.4	0.7475	13.82\\
-2.4	0.797	15.008\\
-2.4	0.8465	16.196\\
-2.4	0.896	17.384\\
-2.4	0.9455	18.572\\
-2.4	0.995	19.76\\
-2.1	0.005	2.205\\
-2.1	0.0545	2.205\\
-2.1	0.104	2.205\\
-2.1	0.1535	2.205\\
-2.1	0.203	2.205\\
-2.1	0.2525	2.205\\
-2.1	0.302	3.047\\
-2.1	0.3515	4.0865\\
-2.1	0.401	5.126\\
-2.1	0.4505	6.1655\\
-2.1	0.5	7.205\\
-2.1	0.5495	8.2445\\
-2.1	0.599	9.284\\
-2.1	0.6485	10.3235\\
-2.1	0.698	11.363\\
-2.1	0.7475	12.4025\\
-2.1	0.797	13.442\\
-2.1	0.8465	14.4815\\
-2.1	0.896	15.521\\
-2.1	0.9455	16.5605\\
-2.1	0.995	17.6\\
-1.8	0.005	1.62\\
-1.8	0.0545	1.62\\
-1.8	0.104	1.62\\
-1.8	0.1535	1.62\\
-1.8	0.203	1.62\\
-1.8	0.2525	2.165\\
-1.8	0.302	3.056\\
-1.8	0.3515	3.947\\
-1.8	0.401	4.838\\
-1.8	0.4505	5.729\\
-1.8	0.5	6.62\\
-1.8	0.5495	7.511\\
-1.8	0.599	8.402\\
-1.8	0.6485	9.293\\
-1.8	0.698	10.184\\
-1.8	0.7475	11.075\\
-1.8	0.797	11.966\\
-1.8	0.8465	12.857\\
-1.8	0.896	13.748\\
-1.8	0.9455	14.639\\
-1.8	0.995	15.53\\
-1.5	0.005	1.125\\
-1.5	0.0545	1.125\\
-1.5	0.104	1.125\\
-1.5	0.1535	1.125\\
-1.5	0.203	1.67\\
-1.5	0.2525	2.4125\\
-1.5	0.302	3.155\\
-1.5	0.3515	3.8975\\
-1.5	0.401	4.64\\
-1.5	0.4505	5.3825\\
-1.5	0.5	6.125\\
-1.5	0.5495	6.8675\\
-1.5	0.599	7.61\\
-1.5	0.6485	8.3525\\
-1.5	0.698	9.095\\
-1.5	0.7475	9.8375\\
-1.5	0.797	10.58\\
-1.5	0.8465	11.3225\\
-1.5	0.896	12.065\\
-1.5	0.9455	12.8075\\
-1.5	0.995	13.55\\
-1.2	0.005	0.72\\
-1.2	0.0545	0.72\\
-1.2	0.104	0.968\\
-1.2	0.1535	1.562\\
-1.2	0.203	2.156\\
-1.2	0.2525	2.75\\
-1.2	0.302	3.344\\
-1.2	0.3515	3.938\\
-1.2	0.401	4.532\\
-1.2	0.4505	5.126\\
-1.2	0.5	5.72\\
-1.2	0.5495	6.314\\
-1.2	0.599	6.908\\
-1.2	0.6485	7.502\\
-1.2	0.698	8.096\\
-1.2	0.7475	8.69\\
-1.2	0.797	9.284\\
-1.2	0.8465	9.878\\
-1.2	0.896	10.472\\
-1.2	0.9455	11.066\\
-1.2	0.995	11.66\\
-0.9	0.005	0.95\\
-0.9	0.0545	1.3955\\
-0.9	0.104	1.841\\
-0.9	0.1535	2.2865\\
-0.9	0.203	2.732\\
-0.9	0.2525	3.1775\\
-0.9	0.302	3.623\\
-0.9	0.3515	4.0685\\
-0.9	0.401	4.514\\
-0.9	0.4505	4.9595\\
-0.9	0.5	5.405\\
-0.9	0.5495	5.8505\\
-0.9	0.599	6.296\\
-0.9	0.6485	6.7415\\
-0.9	0.698	7.187\\
-0.9	0.7475	7.6325\\
-0.9	0.797	8.078\\
-0.9	0.8465	8.5235\\
-0.9	0.896	8.969\\
-0.9	0.9455	9.4145\\
-0.9	0.995	9.86\\
-0.6	0.005	2.21\\
-0.6	0.0545	2.507\\
-0.6	0.104	2.804\\
-0.6	0.1535	3.101\\
-0.6	0.203	3.398\\
-0.6	0.2525	3.695\\
-0.6	0.302	3.992\\
-0.6	0.3515	4.289\\
-0.6	0.401	4.586\\
-0.6	0.4505	4.883\\
-0.6	0.5	5.18\\
-0.6	0.5495	5.477\\
-0.6	0.599	5.774\\
-0.6	0.6485	6.071\\
-0.6	0.698	6.368\\
-0.6	0.7475	6.665\\
-0.6	0.797	6.962\\
-0.6	0.8465	7.259\\
-0.6	0.896	7.556\\
-0.6	0.9455	7.853\\
-0.6	0.995	8.15\\
-0.3	0.005	3.56\\
-0.3	0.0545	3.7085\\
-0.3	0.104	3.857\\
-0.3	0.1535	4.0055\\
-0.3	0.203	4.154\\
-0.3	0.2525	4.3025\\
-0.3	0.302	4.451\\
-0.3	0.3515	4.5995\\
-0.3	0.401	4.748\\
-0.3	0.4505	4.8965\\
-0.3	0.5	5.045\\
-0.3	0.5495	5.1935\\
-0.3	0.599	5.342\\
-0.3	0.6485	5.4905\\
-0.3	0.698	5.639\\
-0.3	0.7475	5.7875\\
-0.3	0.797	5.936\\
-0.3	0.8465	6.0845\\
-0.3	0.896	6.233\\
-0.3	0.9455	6.3815\\
-0.3	0.995	6.53\\
0	0.005	5\\
0	0.0545	5\\
0	0.104	5\\
0	0.1535	5\\
0	0.203	5\\
0	0.2525	5\\
0	0.302	5\\
0	0.3515	5\\
0	0.401	5\\
0	0.4505	5\\
0	0.5	5\\
0	0.5495	5\\
0	0.599	5\\
0	0.6485	5\\
0	0.698	5\\
0	0.7475	5\\
0	0.797	5\\
0	0.8465	5\\
0	0.896	5\\
0	0.9455	5\\
0	0.995	5\\
0.3	0.005	6.53\\
0.3	0.0545	6.3815\\
0.3	0.104	6.233\\
0.3	0.1535	6.0845\\
0.3	0.203	5.936\\
0.3	0.2525	5.7875\\
0.3	0.302	5.639\\
0.3	0.3515	5.4905\\
0.3	0.401	5.342\\
0.3	0.4505	5.1935\\
0.3	0.5	5.045\\
0.3	0.5495	4.8965\\
0.3	0.599	4.748\\
0.3	0.6485	4.5995\\
0.3	0.698	4.451\\
0.3	0.7475	4.3025\\
0.3	0.797	4.154\\
0.3	0.8465	4.0055\\
0.3	0.896	3.857\\
0.3	0.9455	3.7085\\
0.3	0.995	3.56\\
0.6	0.005	8.15\\
0.6	0.0545	7.853\\
0.6	0.104	7.556\\
0.6	0.1535	7.259\\
0.6	0.203	6.962\\
0.6	0.2525	6.665\\
0.6	0.302	6.368\\
0.6	0.3515	6.071\\
0.6	0.401	5.774\\
0.6	0.4505	5.477\\
0.6	0.5	5.18\\
0.6	0.5495	4.883\\
0.6	0.599	4.586\\
0.6	0.6485	4.289\\
0.6	0.698	3.992\\
0.6	0.7475	3.695\\
0.6	0.797	3.398\\
0.6	0.8465	3.101\\
0.6	0.896	2.804\\
0.6	0.9455	2.507\\
0.6	0.995	2.21\\
0.9	0.005	9.86\\
0.9	0.0545	9.4145\\
0.9	0.104	8.969\\
0.9	0.1535	8.5235\\
0.9	0.203	8.078\\
0.9	0.2525	7.6325\\
0.9	0.302	7.187\\
0.9	0.3515	6.7415\\
0.9	0.401	6.296\\
0.9	0.4505	5.8505\\
0.9	0.5	5.405\\
0.9	0.5495	4.9595\\
0.9	0.599	4.514\\
0.9	0.6485	4.0685\\
0.9	0.698	3.623\\
0.9	0.7475	3.1775\\
0.9	0.797	2.732\\
0.9	0.8465	2.2865\\
0.9	0.896	1.841\\
0.9	0.9455	1.3955\\
0.9	0.995	0.95\\
1.2	0.005	11.66\\
1.2	0.0545	11.066\\
1.2	0.104	10.472\\
1.2	0.1535	9.878\\
1.2	0.203	9.284\\
1.2	0.2525	8.69\\
1.2	0.302	8.096\\
1.2	0.3515	7.502\\
1.2	0.401	6.908\\
1.2	0.4505	6.314\\
1.2	0.5	5.72\\
1.2	0.5495	5.126\\
1.2	0.599	4.532\\
1.2	0.6485	3.938\\
1.2	0.698	3.344\\
1.2	0.7475	2.75\\
1.2	0.797	2.156\\
1.2	0.8465	1.562\\
1.2	0.896	0.968\\
1.2	0.9455	0.72\\
1.2	0.995	0.72\\
1.5	0.005	13.55\\
1.5	0.0545	12.8075\\
1.5	0.104	12.065\\
1.5	0.1535	11.3225\\
1.5	0.203	10.58\\
1.5	0.2525	9.8375\\
1.5	0.302	9.095\\
1.5	0.3515	8.3525\\
1.5	0.401	7.61\\
1.5	0.4505	6.8675\\
1.5	0.5	6.125\\
1.5	0.5495	5.3825\\
1.5	0.599	4.64\\
1.5	0.6485	3.8975\\
1.5	0.698	3.155\\
1.5	0.7475	2.4125\\
1.5	0.797	1.67\\
1.5	0.8465	1.125\\
1.5	0.896	1.125\\
1.5	0.9455	1.125\\
1.5	0.995	1.125\\
1.8	0.005	15.53\\
1.8	0.0545	14.639\\
1.8	0.104	13.748\\
1.8	0.1535	12.857\\
1.8	0.203	11.966\\
1.8	0.2525	11.075\\
1.8	0.302	10.184\\
1.8	0.3515	9.293\\
1.8	0.401	8.402\\
1.8	0.4505	7.511\\
1.8	0.5	6.62\\
1.8	0.5495	5.729\\
1.8	0.599	4.838\\
1.8	0.6485	3.947\\
1.8	0.698	3.056\\
1.8	0.7475	2.165\\
1.8	0.797	1.62\\
1.8	0.8465	1.62\\
1.8	0.896	1.62\\
1.8	0.9455	1.62\\
1.8	0.995	1.62\\
2.1	0.005	17.6\\
2.1	0.0545	16.5605\\
2.1	0.104	15.521\\
2.1	0.1535	14.4815\\
2.1	0.203	13.442\\
2.1	0.2525	12.4025\\
2.1	0.302	11.363\\
2.1	0.3515	10.3235\\
2.1	0.401	9.284\\
2.1	0.4505	8.2445\\
2.1	0.5	7.205\\
2.1	0.5495	6.1655\\
2.1	0.599	5.126\\
2.1	0.6485	4.0865\\
2.1	0.698	3.047\\
2.1	0.7475	2.205\\
2.1	0.797	2.205\\
2.1	0.8465	2.205\\
2.1	0.896	2.205\\
2.1	0.9455	2.205\\
2.1	0.995	2.205\\
2.4	0.005	19.76\\
2.4	0.0545	18.572\\
2.4	0.104	17.384\\
2.4	0.1535	16.196\\
2.4	0.203	15.008\\
2.4	0.2525	13.82\\
2.4	0.302	12.632\\
2.4	0.3515	11.444\\
2.4	0.401	10.256\\
2.4	0.4505	9.068\\
2.4	0.5	7.88\\
2.4	0.5495	6.692\\
2.4	0.599	5.504\\
2.4	0.6485	4.316\\
2.4	0.698	3.128\\
2.4	0.7475	2.88\\
2.4	0.797	2.88\\
2.4	0.8465	2.88\\
2.4	0.896	2.88\\
2.4	0.9455	2.88\\
2.4	0.995	2.88\\
2.7	0.005	22.01\\
2.7	0.0545	20.6735\\
2.7	0.104	19.337\\
2.7	0.1535	18.0005\\
2.7	0.203	16.664\\
2.7	0.2525	15.3275\\
2.7	0.302	13.991\\
2.7	0.3515	12.6545\\
2.7	0.401	11.318\\
2.7	0.4505	9.9815\\
2.7	0.5	8.645\\
2.7	0.5495	7.3085\\
2.7	0.599	5.972\\
2.7	0.6485	4.6355\\
2.7	0.698	3.645\\
2.7	0.7475	3.645\\
2.7	0.797	3.645\\
2.7	0.8465	3.645\\
2.7	0.896	3.645\\
2.7	0.9455	3.645\\
2.7	0.995	3.645\\
3	0.005	24.35\\
3	0.0545	22.865\\
3	0.104	21.38\\
3	0.1535	19.895\\
3	0.203	18.41\\
3	0.2525	16.925\\
3	0.302	15.44\\
3	0.3515	13.955\\
3	0.401	12.47\\
3	0.4505	10.985\\
3	0.5	9.5\\
3	0.5495	8.015\\
3	0.599	6.53\\
3	0.6485	5.045\\
3	0.698	4.5\\
3	0.7475	4.5\\
3	0.797	4.5\\
3	0.8465	4.5\\
3	0.896	4.5\\
3	0.9455	4.5\\
3	0.995	4.5\\
};

 \end{axis}
\end{tikzpicture}%
}\hfill
\subfigure[]{
\label{fig:mip-trivial}
%
%
\begin{tikzpicture}
\begin{axis}[%
    width=0.33\textwidth,
    xmin=-3,
    xmax=3,
    tick align=outside,
    xlabel={$\theta$},
    xlabel style={yshift=3ex},
    xmajorgrids,
    ymin=0,
    ymax=1,
    ylabel={y},
    ylabel style={yshift=3ex},
    ymajorgrids,
    zmin=0,
    zmax=24.5,
    zmajorgrids,
    view={-23.5}{30},
    axis x line*=bottom,
    axis y line*=left,
    axis z line*=left,
    colormap={mycolormap}{color(0)=(white) color(1)=(mycolor1)},
]
\addplot3[%
    surf,
    shader=faceted,
    mesh/rows=21
]
table[
    row sep=crcr,
    header=false
] {%
-3	0.005	0\\
-3	0.0545	0\\
-3	0.104	0\\
-3	0.1535	0\\
-3	0.203	0\\
-3	0.2525	0\\
-3	0.302	0\\
-3	0.3515	0\\
-3	0.401	0\\
-3	0.4505	0\\
-3	0.5	0\\
-3	0.5495	0\\
-3	0.599	0\\
-3	0.6485	0\\
-3	0.698	0\\
-3	0.7475	0\\
-3	0.797	0\\
-3	0.8465	0\\
-3	0.896	0\\
-3	0.9455	0\\
-3	0.995	0\\
-2.7	0.005	0\\
-2.7	0.0545	0\\
-2.7	0.104	0\\
-2.7	0.1535	0\\
-2.7	0.203	0\\
-2.7	0.2525	0\\
-2.7	0.302	0\\
-2.7	0.3515	0\\
-2.7	0.401	0\\
-2.7	0.4505	0\\
-2.7	0.5	0\\
-2.7	0.5495	0\\
-2.7	0.599	0\\
-2.7	0.6485	0\\
-2.7	0.698	0\\
-2.7	0.7475	0\\
-2.7	0.797	0\\
-2.7	0.8465	0\\
-2.7	0.896	0\\
-2.7	0.9455	0\\
-2.7	0.995	0\\
-2.4	0.005	0\\
-2.4	0.0545	0\\
-2.4	0.104	0\\
-2.4	0.1535	0\\
-2.4	0.203	0\\
-2.4	0.2525	0\\
-2.4	0.302	0\\
-2.4	0.3515	0\\
-2.4	0.401	0\\
-2.4	0.4505	0\\
-2.4	0.5	0\\
-2.4	0.5495	0\\
-2.4	0.599	0\\
-2.4	0.6485	0\\
-2.4	0.698	0\\
-2.4	0.7475	0\\
-2.4	0.797	0\\
-2.4	0.8465	0\\
-2.4	0.896	0\\
-2.4	0.9455	0\\
-2.4	0.995	0\\
-2.1	0.005	0\\
-2.1	0.0545	0\\
-2.1	0.104	0\\
-2.1	0.1535	0\\
-2.1	0.203	0\\
-2.1	0.2525	0\\
-2.1	0.302	0\\
-2.1	0.3515	0\\
-2.1	0.401	0\\
-2.1	0.4505	0\\
-2.1	0.5	0\\
-2.1	0.5495	0\\
-2.1	0.599	0\\
-2.1	0.6485	0\\
-2.1	0.698	0\\
-2.1	0.7475	0\\
-2.1	0.797	0\\
-2.1	0.8465	0\\
-2.1	0.896	0\\
-2.1	0.9455	0\\
-2.1	0.995	0\\
-1.8	0.005	0\\
-1.8	0.0545	0\\
-1.8	0.104	0\\
-1.8	0.1535	0\\
-1.8	0.203	0\\
-1.8	0.2525	0\\
-1.8	0.302	0\\
-1.8	0.3515	0\\
-1.8	0.401	0\\
-1.8	0.4505	0\\
-1.8	0.5	0\\
-1.8	0.5495	0\\
-1.8	0.599	0\\
-1.8	0.6485	0\\
-1.8	0.698	0\\
-1.8	0.7475	0\\
-1.8	0.797	0\\
-1.8	0.8465	0\\
-1.8	0.896	0\\
-1.8	0.9455	0\\
-1.8	0.995	0\\
-1.5	0.005	0\\
-1.5	0.0545	0\\
-1.5	0.104	0\\
-1.5	0.1535	0\\
-1.5	0.203	0\\
-1.5	0.2525	0\\
-1.5	0.302	0\\
-1.5	0.3515	0\\
-1.5	0.401	0\\
-1.5	0.4505	0\\
-1.5	0.5	0\\
-1.5	0.5495	0\\
-1.5	0.599	0\\
-1.5	0.6485	0\\
-1.5	0.698	0\\
-1.5	0.7475	0\\
-1.5	0.797	0\\
-1.5	0.8465	0\\
-1.5	0.896	0\\
-1.5	0.9455	0\\
-1.5	0.995	0\\
-1.2	0.005	0\\
-1.2	0.0545	0\\
-1.2	0.104	0\\
-1.2	0.1535	0\\
-1.2	0.203	0\\
-1.2	0.2525	0\\
-1.2	0.302	0\\
-1.2	0.3515	0\\
-1.2	0.401	0\\
-1.2	0.4505	0\\
-1.2	0.5	0\\
-1.2	0.5495	0\\
-1.2	0.599	0\\
-1.2	0.6485	0\\
-1.2	0.698	0\\
-1.2	0.7475	0\\
-1.2	0.797	0\\
-1.2	0.8465	0\\
-1.2	0.896	0\\
-1.2	0.9455	0\\
-1.2	0.995	0\\
-0.9	0.005	0\\
-0.9	0.0545	0\\
-0.9	0.104	0\\
-0.9	0.1535	0\\
-0.9	0.203	0\\
-0.9	0.2525	0\\
-0.9	0.302	0\\
-0.9	0.3515	0\\
-0.9	0.401	0\\
-0.9	0.4505	0\\
-0.9	0.5	0\\
-0.9	0.5495	0\\
-0.9	0.599	0\\
-0.9	0.6485	0\\
-0.9	0.698	0\\
-0.9	0.7475	0\\
-0.9	0.797	0\\
-0.9	0.8465	0\\
-0.9	0.896	0\\
-0.9	0.9455	0\\
-0.9	0.995	0\\
-0.6	0.005	0\\
-0.6	0.0545	0\\
-0.6	0.104	0\\
-0.6	0.1535	0\\
-0.6	0.203	0\\
-0.6	0.2525	0\\
-0.6	0.302	0\\
-0.6	0.3515	0\\
-0.6	0.401	0\\
-0.6	0.4505	0\\
-0.6	0.5	0\\
-0.6	0.5495	0\\
-0.6	0.599	0\\
-0.6	0.6485	0\\
-0.6	0.698	0\\
-0.6	0.7475	0\\
-0.6	0.797	0\\
-0.6	0.8465	0\\
-0.6	0.896	0\\
-0.6	0.9455	0\\
-0.6	0.995	0\\
-0.3	0.005	0\\
-0.3	0.0545	0\\
-0.3	0.104	0\\
-0.3	0.1535	0\\
-0.3	0.203	0\\
-0.3	0.2525	0\\
-0.3	0.302	0\\
-0.3	0.3515	0\\
-0.3	0.401	0\\
-0.3	0.4505	0\\
-0.3	0.5	0\\
-0.3	0.5495	0\\
-0.3	0.599	0\\
-0.3	0.6485	0\\
-0.3	0.698	0\\
-0.3	0.7475	0\\
-0.3	0.797	0\\
-0.3	0.8465	0\\
-0.3	0.896	0\\
-0.3	0.9455	0\\
-0.3	0.995	0\\
0	0.005	0\\
0	0.0545	0\\
0	0.104	0\\
0	0.1535	0\\
0	0.203	0\\
0	0.2525	0\\
0	0.302	0\\
0	0.3515	0\\
0	0.401	0\\
0	0.4505	0\\
0	0.5	0\\
0	0.5495	0\\
0	0.599	0\\
0	0.6485	0\\
0	0.698	0\\
0	0.7475	0\\
0	0.797	0\\
0	0.8465	0\\
0	0.896	0\\
0	0.9455	0\\
0	0.995	0\\
0.3	0.005	0\\
0.3	0.0545	0\\
0.3	0.104	0\\
0.3	0.1535	0\\
0.3	0.203	0\\
0.3	0.2525	0\\
0.3	0.302	0\\
0.3	0.3515	0\\
0.3	0.401	0\\
0.3	0.4505	0\\
0.3	0.5	0\\
0.3	0.5495	0\\
0.3	0.599	0\\
0.3	0.6485	0\\
0.3	0.698	0\\
0.3	0.7475	0\\
0.3	0.797	0\\
0.3	0.8465	0\\
0.3	0.896	0\\
0.3	0.9455	0\\
0.3	0.995	0\\
0.6	0.005	0\\
0.6	0.0545	0\\
0.6	0.104	0\\
0.6	0.1535	0\\
0.6	0.203	0\\
0.6	0.2525	0\\
0.6	0.302	0\\
0.6	0.3515	0\\
0.6	0.401	0\\
0.6	0.4505	0\\
0.6	0.5	0\\
0.6	0.5495	0\\
0.6	0.599	0\\
0.6	0.6485	0\\
0.6	0.698	0\\
0.6	0.7475	0\\
0.6	0.797	0\\
0.6	0.8465	0\\
0.6	0.896	0\\
0.6	0.9455	0\\
0.6	0.995	0\\
0.9	0.005	0\\
0.9	0.0545	0\\
0.9	0.104	0\\
0.9	0.1535	0\\
0.9	0.203	0\\
0.9	0.2525	0\\
0.9	0.302	0\\
0.9	0.3515	0\\
0.9	0.401	0\\
0.9	0.4505	0\\
0.9	0.5	0\\
0.9	0.5495	0\\
0.9	0.599	0\\
0.9	0.6485	0\\
0.9	0.698	0\\
0.9	0.7475	0\\
0.9	0.797	0\\
0.9	0.8465	0\\
0.9	0.896	0\\
0.9	0.9455	0\\
0.9	0.995	0\\
1.2	0.005	0\\
1.2	0.0545	0\\
1.2	0.104	0\\
1.2	0.1535	0\\
1.2	0.203	0\\
1.2	0.2525	0\\
1.2	0.302	0\\
1.2	0.3515	0\\
1.2	0.401	0\\
1.2	0.4505	0\\
1.2	0.5	0\\
1.2	0.5495	0\\
1.2	0.599	0\\
1.2	0.6485	0\\
1.2	0.698	0\\
1.2	0.7475	0\\
1.2	0.797	0\\
1.2	0.8465	0\\
1.2	0.896	0\\
1.2	0.9455	0\\
1.2	0.995	0\\
1.5	0.005	0\\
1.5	0.0545	0\\
1.5	0.104	0\\
1.5	0.1535	0\\
1.5	0.203	0\\
1.5	0.2525	0\\
1.5	0.302	0\\
1.5	0.3515	0\\
1.5	0.401	0\\
1.5	0.4505	0\\
1.5	0.5	0\\
1.5	0.5495	0\\
1.5	0.599	0\\
1.5	0.6485	0\\
1.5	0.698	0\\
1.5	0.7475	0\\
1.5	0.797	0\\
1.5	0.8465	0\\
1.5	0.896	0\\
1.5	0.9455	0\\
1.5	0.995	0\\
1.8	0.005	0\\
1.8	0.0545	0\\
1.8	0.104	0\\
1.8	0.1535	0\\
1.8	0.203	0\\
1.8	0.2525	0\\
1.8	0.302	0\\
1.8	0.3515	0\\
1.8	0.401	0\\
1.8	0.4505	0\\
1.8	0.5	0\\
1.8	0.5495	0\\
1.8	0.599	0\\
1.8	0.6485	0\\
1.8	0.698	0\\
1.8	0.7475	0\\
1.8	0.797	0\\
1.8	0.8465	0\\
1.8	0.896	0\\
1.8	0.9455	0\\
1.8	0.995	0\\
2.1	0.005	0\\
2.1	0.0545	0\\
2.1	0.104	0\\
2.1	0.1535	0\\
2.1	0.203	0\\
2.1	0.2525	0\\
2.1	0.302	0\\
2.1	0.3515	0\\
2.1	0.401	0\\
2.1	0.4505	0\\
2.1	0.5	0\\
2.1	0.5495	0\\
2.1	0.599	0\\
2.1	0.6485	0\\
2.1	0.698	0\\
2.1	0.7475	0\\
2.1	0.797	0\\
2.1	0.8465	0\\
2.1	0.896	0\\
2.1	0.9455	0\\
2.1	0.995	0\\
2.4	0.005	0\\
2.4	0.0545	0\\
2.4	0.104	0\\
2.4	0.1535	0\\
2.4	0.203	0\\
2.4	0.2525	0\\
2.4	0.302	0\\
2.4	0.3515	0\\
2.4	0.401	0\\
2.4	0.4505	0\\
2.4	0.5	0\\
2.4	0.5495	0\\
2.4	0.599	0\\
2.4	0.6485	0\\
2.4	0.698	0\\
2.4	0.7475	0\\
2.4	0.797	0\\
2.4	0.8465	0\\
2.4	0.896	0\\
2.4	0.9455	0\\
2.4	0.995	0\\
2.7	0.005	0\\
2.7	0.0545	0\\
2.7	0.104	0\\
2.7	0.1535	0\\
2.7	0.203	0\\
2.7	0.2525	0\\
2.7	0.302	0\\
2.7	0.3515	0\\
2.7	0.401	0\\
2.7	0.4505	0\\
2.7	0.5	0\\
2.7	0.5495	0\\
2.7	0.599	0\\
2.7	0.6485	0\\
2.7	0.698	0\\
2.7	0.7475	0\\
2.7	0.797	0\\
2.7	0.8465	0\\
2.7	0.896	0\\
2.7	0.9455	0\\
2.7	0.995	0\\
3	0.005	0\\
3	0.0545	0\\
3	0.104	0\\
3	0.1535	0\\
3	0.203	0\\
3	0.2525	0\\
3	0.302	0\\
3	0.3515	0\\
3	0.401	0\\
3	0.4505	0\\
3	0.5	0\\
3	0.5495	0\\
3	0.599	0\\
3	0.6485	0\\
3	0.698	0\\
3	0.7475	0\\
3	0.797	0\\
3	0.8465	0\\
3	0.896	0\\
3	0.9455	0\\
3	0.995	0\\
};

\addplot3[solid,line width=1pt]
 table[row sep=crcr] {%
-3	1	24.5\\
-2.7	1	22.145\\
-2.4	1	19.88\\
-2.1	1	17.705\\
-1.8	1	15.62\\
-1.5	1	13.625\\
-1.2	1	11.72\\
-0.9	1	9.905\\
-0.6	1	8.18\\
-0.3	1	6.545\\
0	1	5\\
0.3	1	3.545\\
0.6	1	2.18\\
0.9	1	0.905\\
1.2	1	0.72\\
1.5	1	1.125\\
1.8	1	1.62\\
2.1	1	2.205\\
2.4	1	2.88\\
2.7	1	3.645\\
3	1	4.5\\
};

\addplot3[solid,line width=1pt]
 table[row sep=crcr] {%
-3	0	4.5\\
-2.7	0	3.645\\
-2.4	0	2.88\\
-2.1	0	2.205\\
-1.8	0	1.62\\
-1.5	0	1.125\\
-1.2	0	0.72\\
-0.9	0	0.905\\
-0.6	0	2.18\\
-0.3	0	3.545\\
0	0	5\\
0.3	0	6.545\\
0.6	0	8.18\\
0.9	0	9.905\\
1.2	0	11.72\\
1.5	0	13.625\\
1.8	0	15.62\\
2.1	0	17.705\\
2.4	0	19.88\\
2.7	0	22.145\\
3	0	24.5\\
};

\end{axis}
\end{tikzpicture}%
}\hfill
\subfigure[]{
\label{fig:mip-convex-envelope}
\input{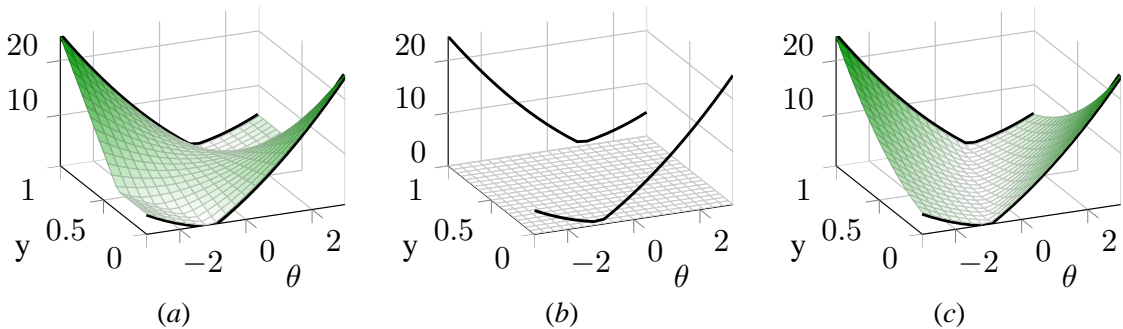}
}
\caption{Depicted above in (a) is the non-convex objective function $\varphi$ 
of the optimization problem
\eqref{eq:problem-main}
for $Y = \{0,1\}$, $\Theta = \mathbb{R}$,
the Hinge loss 
(Tab.~\ref{table:loss-functions})
and 
$\omega(\cdot) = \|\cdot\|_2^2$. 
Its restriction $\phi$ to the feasible set $\Theta \times \{0,1\}$ is depicted in black.
As the goal is to minimize $\varphi$ over the feasible set,
one can replace the values of $\varphi$ for $y \in (0,1)$ without affecting the 
solution, for instance, by zero, as depicted in (b).
In this paper, we characterize the tightest convex extension of $\phi$ to $\Theta \times \conv Y$,
which is depicted in (c).}
\label{fig:ext_description}
\end{figure*} 
We construct,
for the MIP
\eqref{eq:mip-objective}--\eqref{eq:mip-integrality}
whose objective function is non-convex,
MIPs with the same solutions whose objective functions are convex.
Our approach is illustrated in 
Fig.~\ref{fig:ext_description}
and is summarized below.

In Section~\ref{section:relaxation},
we consider the restriction $\phi$ of $\varphi$ to the feasible set $\Theta \times Y$
and characterize the tightest convex extension $\phi^{**}$ of $\phi$ to $\Theta \times \conv\,Y$.
This tightest convex extension $\phi^{**}$ is mostly of theoretical interest as computing its values is NP-hard. 

In Section~\ref{section:characterization}, 
we consider a decomposition of the function $\phi$ into a sum of functions.
By applying our characterization of tightest convex extensions to every function in this sum,
we construct a convex extension $\phi'$ of $\phi$ to $\Theta \times [0,1]^S$ which is not generally tight but whose values can be computed efficiently.
For common loss and regularization functions, we derive a closed form.

For every convex extension $\phi'$ we construct, including $\phi^{**}$,
the MIP
\begin{align}
\inf_{(\theta, y) \in \mathbb{R}^m \times \mathbb{R}^S} \quad
    & \phi'(\theta, y) 
    \label{eq:convex-mip-objective}\\
\textnormal{subject to} \quad
    & A' \theta \leq b' \label{eq:convex-mpi-theta-polytope} \\
    & A y \leq b \label{eq:convex-mip-label-polytope}\\
    & y \in \{0,1\}^S \label{eq:convex-mip-integrality}
\end{align}
has the same solutions as
\eqref{eq:mip-objective}--\eqref{eq:mip-integrality}.
Like 
\eqref{eq:mip-objective}--\eqref{eq:mip-integrality},
it is NP-hard, due to the integrality constraint
\eqref{eq:convex-mip-integrality}.
Unlike
\eqref{eq:mip-objective}--\eqref{eq:mip-integrality},
its objective function and polyhedral relaxation 
\eqref{eq:convex-mip-objective}--\eqref{eq:convex-mip-label-polytope}
are convex. 
Thus, unlike
\eqref{eq:mip-objective}--\eqref{eq:mip-integrality},
it is accessible to a wide range of standard optimization techniques.

\section{Related Work}

\subsection{Convex Extensions}
For large classes of univariate and bivariate functions, 
tightest convex extensions have been characterized by
\citet{tawarmalani2013explicit}
and
\citet{locatelli2014technique}.
Convex envelopes of multivariate functions that are convex in all but one variable have been characterized by
\citet{jach2008convex}.
For functions of the form $f(x,y) = g(x) h(y)$ and with $g$ and $h$ having additional properties,
e.g., $g$ being component-wise concave and submodular, and $h$ being univariate convex,
tightest convex extensions have been characterized by
\cite{khajavirad2012convex, khajavirad2013convex}.
For functions $f: A \times B \to \mathbb{R}$ with $A, B \subseteq \mathbb{R}^n$ 
and $f(a,\cdot)$ being either convex or concave
and $f(\cdot,b)$ being either convex or concave
for any $a \in A$ and $b \in B$,\
tightest convex extensions have been characterized by
\citet{ballerstein2013convex}.
Tightest convex extensions of pseudo-Boolean functions $f: \{0,1\}^n \to \mathbb{R}$
are known as convex closures
\citep{MAL-039}.
Convex closures of submodular functions are Lov\'asz extensions.

The characterization of tightest convex extensions 
of functions $f: \Theta \times Y \to \mathbb{R}$ with $f(\cdot, y)$ convex for all $y \in Y$
that we establish is consistent with results of
\citet{jach2008convex}
for functions $f: \mathbb{R}^n \times \{0,1\} \to \mathbb{R}$.
It extends some results of
\citet{jach2008convex}
to non-differentiable $f(\cdot, 0)$ and $f(\cdot, 1)$.

\subsection{Regularized Empirical Risk Minimization with Constrained Labels}
Regularized empirical risk minimization with constrained labels
has been studied intensively in the special case of semi-supervised learning for 01-classification.
Algorithms that find feasible solutions efficiently are due to
\citet{joachims-1999,joachims-2003,chapelle-2005-semi,chapelle-2006-continuation,chapelle-2006-branch,sindhwani2006deterministic}.
A branch-and-bound algorithm that solves the problem to optimality was suggested by
\citet{vapnik1974theory}
and has been implemented and applied to data by
\citet{chapelle-2006-branch,chapelle-2008-optimization},
with the result that optimal solutions generalize better typically than feasible solutions found by approximate algorithms.

An approach to the problem by convex optimization, specifically, by a semi-definite relaxation of the dual problem, was proposed by
\citet{debie-2006}.
Similar relaxations have been studied in the context of semi-supervised learning for multi-label classification
\citep{xu-2005-multi-class,guo2011adaptive},
correlation clustering
\citep{xu-2005-clustering,zhang2009maximum}
and latent variable estimation
\citep{guo2008convex}.
For maximum margin clustering, a convex relaxation tighter than the SDP relaxation is constructed by \citet{li-2009}.
For multi-label classification with a softmax loss function, a tight SDP relaxation is proposed by \cite{joulin2012convex}.

\newpage
\section{Tightest Convex Extensions}
\label{section:relaxation}
In this section, we consider the restriction $\phi$ of $\varphi$ to the feasible set $\Theta \times Y$,
i.e., the function 
\begin{align}
\phi: 
    \quad \Theta \times Y \to \mathbb{R}_0^+: 
    \quad (\theta,y) \mapsto \varphi(\theta, y)
\enspace .
\end{align}
We characterize the tightest convex extension $\phi^{**}$ of $\phi$ to $\Theta \times \conv\,Y$
in Theorem~\ref{theorem:main}.
\begin{definition}
\citep{tawarmalani2002convexification}
For any $n \in \mathbb{N}$, any $A \subseteq \mathbb{R}^n$ and any $\phi: A \to \mathbb{R}$, 
a function $\phi': \conv\,A \to \mathbb{R}$ is called a \emph{convex extension} of $\phi$ 
iff $\phi'$ is convex and $\forall a \in A: \phi'(a) = \phi(a)$.
Moreover, a function $\phi^{**}: \conv\,A \to \mathbb{R}$ is called the \emph{tightest convex extension} of $\phi$ 
iff $\phi^{**}$ is a convex extension of $\phi$ and, 
for every convex extension $\phi'$ of $\phi$ 
and for all $a \in \conv\,A$: $\phi'(a) \leq \phi^{**}(a)$.
\end{definition}

\begin{lemma}
A (tightest) convex extension $\phi^{**}$ of $\phi$ exists.
\end{lemma}

\begin{theorem} 
\label{theorem:main} 
For every finite $S \neq \emptyset$,
every $Y \subseteq \{0,1\}^S$ with $|Y| > |S|$,
every $m \in \mathbb{N}$, 
every convex $\Theta \subseteq \mathbb{R}^m$
and every $\phi: \Theta \times Y \to \mathbb{R}_0^+$ such that $\phi(\cdot, y)$ is convex for every $y \in Y$,
the tightest convex extension
$\phi^{**}: \Theta \times \conv\,Y \to \mathbb{R}_0^+$ of $\phi$ 
is such that for all $(\theta,y) \in \Theta \times \conv\,Y$:
\begin{align}
\phi^{**}(\theta, y)
    = \min_{Y' \in \mathcal{Y}(y)} 
        \inf_{\theta': Y' \to \Theta} \Bigg\{ 
            \sum_{y' \in Y'} \lambda_{y Y'}(y') \, \phi(\theta'(y'), y')
            \ \Bigg|
            \sum_{y' \in Y'} \lambda_{y Y'}(y') \, \theta'(y') = \theta
        \Bigg\}
\label{eq:main-theorem-1}
\end{align}
where
\begin{align}
\mathcal{Y}(y) := \left\{
    Y' \in \tbinom{Y}{|S| + 1}
    \ \middle|\  
    y \in \conv\,Y'
\right\}
\end{align}
is the set of $(|S| + 1)$-elementary subsets $Y'$ of $Y$ having $y$ in their convex hull,
and, for every $y \in \conv\,Y$ and every $Y' \in \mathcal{Y}(y)$: 
$\lambda_{y Y'}: Y' \to \mathbb{R}_0^+$ are the coefficients in the convex combination of $y$ in $Y'$, i.e.
\begin{align}
\sum_{y' \in Y'} \lambda_{y Y'}(y') \, y' & = y \\
\sum_{y' \in Y'} \lambda_{y Y'}(y') & = 1
\enspace .
\end{align}
\end{theorem}
Remarks are in order:
\begin{itemize}
\itemsep0ex
\item $|\mathcal{Y}(y)|$ can be strongly exponential in $|S|$.
This is expected, given that the complexity of computing the value of a convex extension is NP-hard. 
\item Given a polynomial-time oracle for the outer minimization in
\eqref{eq:main-theorem-1},
the tightest convex extension can be computed in polynomial time.
\item Conceivable is a polynomial-time oracle for constraining $\mathcal{Y}(y)$ to a strict subset without compromising optimality in 
\eqref{eq:main-theorem-1}.
\item An upper bound on the tightest convex extension can be obtained by considering any subset of $\mathcal{Y}(y)$.
\end{itemize}

\section{Efficient Convex Extensions} 
\label{section:characterization}
\begin{figure*}
\subfigure[]{
\label{fig:L1-nonconvex}
%
%
\begin{tikzpicture}

\begin{axis}[%
width=\smallplotwidth,
xmin=-3,
xmax=3,
tick align=outside,
xlabel={$\theta$},
    xlabel style={yshift=3ex},
xmajorgrids,
ymin=0,
ymax=1,
ylabel={y},
    ylabel style={yshift=3ex},
ymajorgrids,
zmin=1.1,
zmax=11,
zmajorgrids,
view={-18.5}{34},
axis x line*=bottom,
axis y line*=left,
axis z line*=left,
colormap={mycolormap}{color(0)=(white) color(1)=(mycolor1)}
]

\addplot3[%
    surf,
    shader=faceted,
    mesh/rows=21
]
table[row sep=crcr,header=false] {%
-3	0.01	3\\
-3	0.059	3\\
-3	0.108	3\\
-3	0.157	3\\
-3	0.206	3\\
-3	0.255	3\\
-3	0.304	3\\
-3	0.353	3.236\\
-3	0.402	3.824\\
-3	0.451	4.412\\
-3	0.5	5\\
-3	0.549	5.588\\
-3	0.598	6.176\\
-3	0.647	6.764\\
-3	0.696	7.352\\
-3	0.745	7.94\\
-3	0.794	8.528\\
-3	0.843	9.116\\
-3	0.892	9.704\\
-3	0.941	10.292\\
-3	0.99	10.88\\
-2.7	0.01	2.7\\
-2.7	0.059	2.7\\
-2.7	0.108	2.7\\
-2.7	0.157	2.7\\
-2.7	0.206	2.7\\
-2.7	0.255	2.7\\
-2.7	0.304	2.7\\
-2.7	0.353	3.1124\\
-2.7	0.402	3.6416\\
-2.7	0.451	4.1708\\
-2.7	0.5	4.7\\
-2.7	0.549	5.2292\\
-2.7	0.598	5.7584\\
-2.7	0.647	6.2876\\
-2.7	0.696	6.8168\\
-2.7	0.745	7.346\\
-2.7	0.794	7.8752\\
-2.7	0.843	8.4044\\
-2.7	0.892	8.9336\\
-2.7	0.941	9.4628\\
-2.7	0.99	9.992\\
-2.4	0.01	2.4\\
-2.4	0.059	2.4\\
-2.4	0.108	2.4\\
-2.4	0.157	2.4\\
-2.4	0.206	2.4\\
-2.4	0.255	2.4\\
-2.4	0.304	2.5184\\
-2.4	0.353	2.9888\\
-2.4	0.402	3.4592\\
-2.4	0.451	3.9296\\
-2.4	0.5	4.4\\
-2.4	0.549	4.8704\\
-2.4	0.598	5.3408\\
-2.4	0.647	5.8112\\
-2.4	0.696	6.2816\\
-2.4	0.745	6.752\\
-2.4	0.794	7.2224\\
-2.4	0.843	7.6928\\
-2.4	0.892	8.1632\\
-2.4	0.941	8.6336\\
-2.4	0.99	9.104\\
-2.1	0.01	2.1\\
-2.1	0.059	2.1\\
-2.1	0.108	2.1\\
-2.1	0.157	2.1\\
-2.1	0.206	2.1\\
-2.1	0.255	2.1\\
-2.1	0.304	2.4536\\
-2.1	0.353	2.8652\\
-2.1	0.402	3.2768\\
-2.1	0.451	3.6884\\
-2.1	0.5	4.1\\
-2.1	0.549	4.5116\\
-2.1	0.598	4.9232\\
-2.1	0.647	5.3348\\
-2.1	0.696	5.7464\\
-2.1	0.745	6.158\\
-2.1	0.794	6.5696\\
-2.1	0.843	6.9812\\
-2.1	0.892	7.3928\\
-2.1	0.941	7.8044\\
-2.1	0.99	8.216\\
-1.8	0.01	1.8\\
-1.8	0.059	1.8\\
-1.8	0.108	1.8\\
-1.8	0.157	1.8\\
-1.8	0.206	1.8\\
-1.8	0.255	2.036\\
-1.8	0.304	2.3888\\
-1.8	0.353	2.7416\\
-1.8	0.402	3.0944\\
-1.8	0.451	3.4472\\
-1.8	0.5	3.8\\
-1.8	0.549	4.1528\\
-1.8	0.598	4.5056\\
-1.8	0.647	4.8584\\
-1.8	0.696	5.2112\\
-1.8	0.745	5.564\\
-1.8	0.794	5.9168\\
-1.8	0.843	6.2696\\
-1.8	0.892	6.6224\\
-1.8	0.941	6.9752\\
-1.8	0.99	7.328\\
-1.5	0.01	1.5\\
-1.5	0.059	1.5\\
-1.5	0.108	1.5\\
-1.5	0.157	1.5\\
-1.5	0.206	1.736\\
-1.5	0.255	2.03\\
-1.5	0.304	2.324\\
-1.5	0.353	2.618\\
-1.5	0.402	2.912\\
-1.5	0.451	3.206\\
-1.5	0.5	3.5\\
-1.5	0.549	3.794\\
-1.5	0.598	4.088\\
-1.5	0.647	4.382\\
-1.5	0.696	4.676\\
-1.5	0.745	4.97\\
-1.5	0.794	5.264\\
-1.5	0.843	5.558\\
-1.5	0.892	5.852\\
-1.5	0.941	6.146\\
-1.5	0.99	6.44\\
-1.2	0.01	1.2\\
-1.2	0.059	1.2\\
-1.2	0.108	1.3184\\
-1.2	0.157	1.5536\\
-1.2	0.206	1.7888\\
-1.2	0.255	2.024\\
-1.2	0.304	2.2592\\
-1.2	0.353	2.4944\\
-1.2	0.402	2.7296\\
-1.2	0.451	2.9648\\
-1.2	0.5	3.2\\
-1.2	0.549	3.4352\\
-1.2	0.598	3.6704\\
-1.2	0.647	3.9056\\
-1.2	0.696	4.1408\\
-1.2	0.745	4.376\\
-1.2	0.794	4.6112\\
-1.2	0.843	4.8464\\
-1.2	0.892	5.0816\\
-1.2	0.941	5.3168\\
-1.2	0.99	5.552\\
-0.9	0.01	1.136\\
-0.9	0.059	1.3124\\
-0.9	0.108	1.4888\\
-0.9	0.157	1.6652\\
-0.9	0.206	1.8416\\
-0.9	0.255	2.018\\
-0.9	0.304	2.1944\\
-0.9	0.353	2.3708\\
-0.9	0.402	2.5472\\
-0.9	0.451	2.7236\\
-0.9	0.5	2.9\\
-0.9	0.549	3.0764\\
-0.9	0.598	3.2528\\
-0.9	0.647	3.4292\\
-0.9	0.696	3.6056\\
-0.9	0.745	3.782\\
-0.9	0.794	3.9584\\
-0.9	0.843	4.1348\\
-0.9	0.892	4.3112\\
-0.9	0.941	4.4876\\
-0.9	0.99	4.664\\
-0.6	0.01	1.424\\
-0.6	0.059	1.5416\\
-0.6	0.108	1.6592\\
-0.6	0.157	1.7768\\
-0.6	0.206	1.8944\\
-0.6	0.255	2.012\\
-0.6	0.304	2.1296\\
-0.6	0.353	2.2472\\
-0.6	0.402	2.3648\\
-0.6	0.451	2.4824\\
-0.6	0.5	2.6\\
-0.6	0.549	2.7176\\
-0.6	0.598	2.8352\\
-0.6	0.647	2.9528\\
-0.6	0.696	3.0704\\
-0.6	0.745	3.188\\
-0.6	0.794	3.3056\\
-0.6	0.843	3.4232\\
-0.6	0.892	3.5408\\
-0.6	0.941	3.6584\\
-0.6	0.99	3.776\\
-0.3	0.01	1.712\\
-0.3	0.059	1.7708\\
-0.3	0.108	1.8296\\
-0.3	0.157	1.8884\\
-0.3	0.206	1.9472\\
-0.3	0.255	2.006\\
-0.3	0.304	2.0648\\
-0.3	0.353	2.1236\\
-0.3	0.402	2.1824\\
-0.3	0.451	2.2412\\
-0.3	0.5	2.3\\
-0.3	0.549	2.3588\\
-0.3	0.598	2.4176\\
-0.3	0.647	2.4764\\
-0.3	0.696	2.5352\\
-0.3	0.745	2.594\\
-0.3	0.794	2.6528\\
-0.3	0.843	2.7116\\
-0.3	0.892	2.7704\\
-0.3	0.941	2.8292\\
-0.3	0.99	2.888\\
0	0.01	2\\
0	0.059	2\\
0	0.108	2\\
0	0.157	2\\
0	0.206	2\\
0	0.255	2\\
0	0.304	2\\
0	0.353	2\\
0	0.402	2\\
0	0.451	2\\
0	0.5	2\\
0	0.549	2\\
0	0.598	2\\
0	0.647	2\\
0	0.696	2\\
0	0.745	2\\
0	0.794	2\\
0	0.843	2\\
0	0.892	2\\
0	0.941	2\\
0	0.99	2\\
0.3	0.01	2.888\\
0.3	0.059	2.8292\\
0.3	0.108	2.7704\\
0.3	0.157	2.7116\\
0.3	0.206	2.6528\\
0.3	0.255	2.594\\
0.3	0.304	2.5352\\
0.3	0.353	2.4764\\
0.3	0.402	2.4176\\
0.3	0.451	2.3588\\
0.3	0.5	2.3\\
0.3	0.549	2.2412\\
0.3	0.598	2.1824\\
0.3	0.647	2.1236\\
0.3	0.696	2.0648\\
0.3	0.745	2.006\\
0.3	0.794	1.9472\\
0.3	0.843	1.8884\\
0.3	0.892	1.8296\\
0.3	0.941	1.7708\\
0.3	0.99	1.712\\
0.6	0.01	3.776\\
0.6	0.059	3.6584\\
0.6	0.108	3.5408\\
0.6	0.157	3.4232\\
0.6	0.206	3.3056\\
0.6	0.255	3.188\\
0.6	0.304	3.0704\\
0.6	0.353	2.9528\\
0.6	0.402	2.8352\\
0.6	0.451	2.7176\\
0.6	0.5	2.6\\
0.6	0.549	2.4824\\
0.6	0.598	2.3648\\
0.6	0.647	2.2472\\
0.6	0.696	2.1296\\
0.6	0.745	2.012\\
0.6	0.794	1.8944\\
0.6	0.843	1.7768\\
0.6	0.892	1.6592\\
0.6	0.941	1.5416\\
0.6	0.99	1.424\\
0.9	0.01	4.664\\
0.9	0.059	4.4876\\
0.9	0.108	4.3112\\
0.9	0.157	4.1348\\
0.9	0.206	3.9584\\
0.9	0.255	3.782\\
0.9	0.304	3.6056\\
0.9	0.353	3.4292\\
0.9	0.402	3.2528\\
0.9	0.451	3.0764\\
0.9	0.5	2.9\\
0.9	0.549	2.7236\\
0.9	0.598	2.5472\\
0.9	0.647	2.3708\\
0.9	0.696	2.1944\\
0.9	0.745	2.018\\
0.9	0.794	1.8416\\
0.9	0.843	1.6652\\
0.9	0.892	1.4888\\
0.9	0.941	1.3124\\
0.9	0.99	1.136\\
1.2	0.01	5.552\\
1.2	0.059	5.3168\\
1.2	0.108	5.0816\\
1.2	0.157	4.8464\\
1.2	0.206	4.6112\\
1.2	0.255	4.376\\
1.2	0.304	4.1408\\
1.2	0.353	3.9056\\
1.2	0.402	3.6704\\
1.2	0.451	3.4352\\
1.2	0.5	3.2\\
1.2	0.549	2.9648\\
1.2	0.598	2.7296\\
1.2	0.647	2.4944\\
1.2	0.696	2.2592\\
1.2	0.745	2.024\\
1.2	0.794	1.7888\\
1.2	0.843	1.5536\\
1.2	0.892	1.3184\\
1.2	0.941	1.2\\
1.2	0.99	1.2\\
1.5	0.01	6.44\\
1.5	0.059	6.146\\
1.5	0.108	5.852\\
1.5	0.157	5.558\\
1.5	0.206	5.264\\
1.5	0.255	4.97\\
1.5	0.304	4.676\\
1.5	0.353	4.382\\
1.5	0.402	4.088\\
1.5	0.451	3.794\\
1.5	0.5	3.5\\
1.5	0.549	3.206\\
1.5	0.598	2.912\\
1.5	0.647	2.618\\
1.5	0.696	2.324\\
1.5	0.745	2.03\\
1.5	0.794	1.736\\
1.5	0.843	1.5\\
1.5	0.892	1.5\\
1.5	0.941	1.5\\
1.5	0.99	1.5\\
1.8	0.01	7.328\\
1.8	0.059	6.9752\\
1.8	0.108	6.6224\\
1.8	0.157	6.2696\\
1.8	0.206	5.9168\\
1.8	0.255	5.564\\
1.8	0.304	5.2112\\
1.8	0.353	4.8584\\
1.8	0.402	4.5056\\
1.8	0.451	4.1528\\
1.8	0.5	3.8\\
1.8	0.549	3.4472\\
1.8	0.598	3.0944\\
1.8	0.647	2.7416\\
1.8	0.696	2.3888\\
1.8	0.745	2.036\\
1.8	0.794	1.8\\
1.8	0.843	1.8\\
1.8	0.892	1.8\\
1.8	0.941	1.8\\
1.8	0.99	1.8\\
2.1	0.01	8.216\\
2.1	0.059	7.8044\\
2.1	0.108	7.3928\\
2.1	0.157	6.9812\\
2.1	0.206	6.5696\\
2.1	0.255	6.158\\
2.1	0.304	5.7464\\
2.1	0.353	5.3348\\
2.1	0.402	4.9232\\
2.1	0.451	4.5116\\
2.1	0.5	4.1\\
2.1	0.549	3.6884\\
2.1	0.598	3.2768\\
2.1	0.647	2.8652\\
2.1	0.696	2.4536\\
2.1	0.745	2.1\\
2.1	0.794	2.1\\
2.1	0.843	2.1\\
2.1	0.892	2.1\\
2.1	0.941	2.1\\
2.1	0.99	2.1\\
2.4	0.01	9.104\\
2.4	0.059	8.6336\\
2.4	0.108	8.1632\\
2.4	0.157	7.6928\\
2.4	0.206	7.2224\\
2.4	0.255	6.752\\
2.4	0.304	6.2816\\
2.4	0.353	5.8112\\
2.4	0.402	5.3408\\
2.4	0.451	4.8704\\
2.4	0.5	4.4\\
2.4	0.549	3.9296\\
2.4	0.598	3.4592\\
2.4	0.647	2.9888\\
2.4	0.696	2.5184\\
2.4	0.745	2.4\\
2.4	0.794	2.4\\
2.4	0.843	2.4\\
2.4	0.892	2.4\\
2.4	0.941	2.4\\
2.4	0.99	2.4\\
2.7	0.01	9.992\\
2.7	0.059	9.4628\\
2.7	0.108	8.9336\\
2.7	0.157	8.4044\\
2.7	0.206	7.8752\\
2.7	0.255	7.346\\
2.7	0.304	6.8168\\
2.7	0.353	6.2876\\
2.7	0.402	5.7584\\
2.7	0.451	5.2292\\
2.7	0.5	4.7\\
2.7	0.549	4.1708\\
2.7	0.598	3.6416\\
2.7	0.647	3.1124\\
2.7	0.696	2.7\\
2.7	0.745	2.7\\
2.7	0.794	2.7\\
2.7	0.843	2.7\\
2.7	0.892	2.7\\
2.7	0.941	2.7\\
2.7	0.99	2.7\\
3	0.01	10.88\\
3	0.059	10.292\\
3	0.108	9.704\\
3	0.157	9.116\\
3	0.206	8.528\\
3	0.255	7.94\\
3	0.304	7.352\\
3	0.353	6.764\\
3	0.402	6.176\\
3	0.451	5.588\\
3	0.5	5\\
3	0.549	4.412\\
3	0.598	3.824\\
3	0.647	3.236\\
3	0.696	3\\
3	0.745	3\\
3	0.794	3\\
3	0.843	3\\
3	0.892	3\\
3	0.941	3\\
3	0.99	3\\
};

\addplot3[solid]
 table[row sep=crcr] {%
-3	0	3\\
-2.7	0	2.7\\
-2.4	0	2.4\\
-2.1	0	2.1\\
-1.8	0	1.8\\
-1.5	0	1.5\\
-1.2	0	1.2\\
-0.9	0	1.1\\
-0.6	0	1.4\\
-0.3	0	1.7\\
0	0	2\\
0.3	0	2.9\\
0.6	0	3.8\\
0.9	0	4.7\\
1.2	0	5.6\\
1.5	0	6.5\\
1.8	0	7.4\\
2.1	0	8.3\\
2.4	0	9.2\\
2.7	0	10.1\\
3	0	11\\
};

\addplot3[solid]
 table[row sep=crcr] {%
-3	1	11\\
-2.7	1	10.1\\
-2.4	1	9.2\\
-2.1	1	8.3\\
-1.8	1	7.4\\
-1.5	1	6.5\\
-1.2	1	5.6\\
-0.9	1	4.7\\
-0.6	1	3.8\\
-0.3	1	2.9\\
0	1	2\\
0.3	1	1.7\\
0.6	1	1.4\\
0.9	1	1.1\\
1.2	1	1.2\\
1.5	1	1.5\\
1.8	1	1.8\\
2.1	1	2.1\\
2.4	1	2.4\\
2.7	1	2.7\\
3	1	3\\
};
 \end{axis}
\end{tikzpicture}%
}\hfill
\subfigure[]{
\label{fig:L1-discontinious}
%
%
\begin{tikzpicture}

\begin{axis}[%
width=\smallplotwidth,
xmin=-3,
xmax=3,
tick align=outside,
xlabel={$\theta$},
    xlabel style={yshift=3ex},
xmajorgrids,
ymin=0,
ymax=1,
ylabel={y},
    ylabel style={yshift=3ex},
ymajorgrids,
zmin=1.00000000160102,
zmax=11,
zmajorgrids,
view={-18.5}{34},
axis x line*=bottom,
axis y line*=left,
axis z line*=left,
colormap={mycolormap}{color(0)=(white) color(1)=(mycolor1)}
]

\addplot3[%
    surf,
    shader=faceted,
    mesh/rows=21
]
table[row sep=crcr,header=false] {%
-3	0.01	3.02\\
-3	0.059	3.118\\
-3	0.108	3.216\\
-3	0.157	3.314\\
-3	0.206	3.412\\
-3	0.255	3.51\\
-3	0.304	3.608\\
-3	0.353	3.706\\
-3	0.402	3.804\\
-3	0.451	3.902\\
-3	0.5	4\\
-3	0.549	4.098\\
-3	0.598	4.196\\
-3	0.647	4.294\\
-3	0.696	4.392\\
-3	0.745	4.49\\
-3	0.794	4.588\\
-3	0.843	4.686\\
-3	0.892	4.784\\
-3	0.941	4.882\\
-3	0.99	4.98\\
-2.7	0.01	2.72\\
-2.7	0.059	2.818\\
-2.7	0.108	2.916\\
-2.7	0.157	3.014\\
-2.7	0.206	3.112\\
-2.7	0.255	3.21\\
-2.7	0.304	3.308\\
-2.7	0.353	3.406\\
-2.7	0.402	3.504\\
-2.7	0.451	3.602\\
-2.7	0.5	3.7\\
-2.7	0.549	3.798\\
-2.7	0.598	3.896\\
-2.7	0.647	3.994\\
-2.7	0.696	4.092\\
-2.7	0.745	4.19\\
-2.7	0.794	4.288\\
-2.7	0.843	4.386\\
-2.7	0.892	4.484\\
-2.7	0.941	4.582\\
-2.7	0.99	4.68\\
-2.4	0.01	2.42\\
-2.4	0.059	2.518\\
-2.4	0.108	2.616\\
-2.4	0.157	2.714\\
-2.4	0.206	2.812\\
-2.4	0.255	2.91\\
-2.4	0.304	3.008\\
-2.4	0.353	3.106\\
-2.4	0.402	3.204\\
-2.4	0.451	3.302\\
-2.4	0.5	3.4\\
-2.4	0.549	3.498\\
-2.4	0.598	3.596\\
-2.4	0.647	3.694\\
-2.4	0.696	3.792\\
-2.4	0.745	3.89\\
-2.4	0.794	3.988\\
-2.4	0.843	4.086\\
-2.4	0.892	4.184\\
-2.4	0.941	4.282\\
-2.4	0.99	4.38\\
-2.1	0.01	2.12\\
-2.1	0.059	2.218\\
-2.1	0.108	2.316\\
-2.1	0.157	2.414\\
-2.1	0.206	2.512\\
-2.1	0.255	2.61\\
-2.1	0.304	2.708\\
-2.1	0.353	2.806\\
-2.1	0.402	2.904\\
-2.1	0.451	3.002\\
-2.1	0.5	3.1\\
-2.1	0.549	3.198\\
-2.1	0.598	3.296\\
-2.1	0.647	3.394\\
-2.1	0.696	3.492\\
-2.1	0.745	3.59\\
-2.1	0.794	3.688\\
-2.1	0.843	3.786\\
-2.1	0.892	3.884\\
-2.1	0.941	3.982\\
-2.1	0.99	4.08\\
-1.8	0.01	1.82\\
-1.8	0.059	1.918\\
-1.8	0.108	2.016\\
-1.8	0.157	2.114\\
-1.8	0.206	2.212\\
-1.8	0.255	2.31\\
-1.8	0.304	2.408\\
-1.8	0.353	2.506\\
-1.8	0.402	2.604\\
-1.8	0.451	2.702\\
-1.8	0.5	2.8\\
-1.8	0.549	2.898\\
-1.8	0.598	2.996\\
-1.8	0.647	3.094\\
-1.8	0.696	3.192\\
-1.8	0.745	3.29\\
-1.8	0.794	3.388\\
-1.8	0.843	3.486\\
-1.8	0.892	3.584\\
-1.8	0.941	3.682\\
-1.8	0.99	3.78\\
-1.5	0.01	1.52\\
-1.5	0.059	1.618\\
-1.5	0.108	1.716\\
-1.5	0.157	1.814\\
-1.5	0.206	1.912\\
-1.5	0.255	2.01\\
-1.5	0.304	2.108\\
-1.5	0.353	2.206\\
-1.5	0.402	2.304\\
-1.5	0.451	2.402\\
-1.5	0.5	2.5\\
-1.5	0.549	2.598\\
-1.5	0.598	2.696\\
-1.5	0.647	2.794\\
-1.5	0.696	2.892\\
-1.5	0.745	2.99\\
-1.5	0.794	3.088\\
-1.5	0.843	3.186\\
-1.5	0.892	3.284\\
-1.5	0.941	3.382\\
-1.5	0.99	3.48\\
-1.2	0.01	1.22\\
-1.2	0.059	1.318\\
-1.2	0.108	1.416\\
-1.2	0.157	1.514\\
-1.2	0.206	1.612\\
-1.2	0.255	1.71\\
-1.2	0.304	1.808\\
-1.2	0.353	1.906\\
-1.2	0.402	2.004\\
-1.2	0.451	2.102\\
-1.2	0.5	2.2\\
-1.2	0.549	2.298\\
-1.2	0.598	2.396\\
-1.2	0.647	2.494\\
-1.2	0.696	2.592\\
-1.2	0.745	2.69\\
-1.2	0.794	2.788\\
-1.2	0.843	2.886\\
-1.2	0.892	2.984\\
-1.2	0.941	3.082\\
-1.2	0.99	3.18\\
-0.9	0.01	1.08\\
-0.9	0.059	1.018\\
-0.9	0.108	1.116\\
-0.9	0.157	1.214\\
-0.9	0.206	1.312\\
-0.9	0.255	1.41\\
-0.9	0.304	1.508\\
-0.9	0.353	1.606\\
-0.9	0.402	1.704\\
-0.9	0.451	1.802\\
-0.9	0.5	1.9\\
-0.9	0.549	1.998\\
-0.9	0.598	2.096\\
-0.9	0.647	2.194\\
-0.9	0.696	2.292\\
-0.9	0.745	2.39\\
-0.9	0.794	2.488\\
-0.9	0.843	2.586\\
-0.9	0.892	2.684\\
-0.9	0.941	2.782\\
-0.9	0.99	2.88\\
-0.6	0.01	1.38\\
-0.6	0.059	1.282\\
-0.6	0.108	1.184\\
-0.6	0.157	1.086\\
-0.6	0.206	1.012\\
-0.6	0.255	1.11\\
-0.6	0.304	1.208\\
-0.6	0.353	1.306\\
-0.6	0.402	1.404\\
-0.6	0.451	1.502\\
-0.6	0.5	1.6\\
-0.6	0.549	1.698\\
-0.6	0.598	1.796\\
-0.6	0.647	1.894\\
-0.6	0.696	1.992\\
-0.6	0.745	2.09\\
-0.6	0.794	2.188\\
-0.6	0.843	2.286\\
-0.6	0.892	2.384\\
-0.6	0.941	2.482\\
-0.6	0.99	2.58\\
-0.3	0.01	1.68\\
-0.3	0.059	1.582\\
-0.3	0.108	1.484\\
-0.3	0.157	1.386\\
-0.3	0.206	1.288\\
-0.3	0.255	1.19\\
-0.3	0.304	1.092\\
-0.3	0.353	1.006\\
-0.3	0.402	1.104\\
-0.3	0.451	1.202\\
-0.3	0.5	1.3\\
-0.3	0.549	1.398\\
-0.3	0.598	1.496\\
-0.3	0.647	1.594\\
-0.3	0.696	1.692\\
-0.3	0.745	1.79\\
-0.3	0.794	1.888\\
-0.3	0.843	1.986\\
-0.3	0.892	2.084\\
-0.3	0.941	2.182\\
-0.3	0.99	2.28\\
0	0.01	1.98\\
0	0.059	1.882\\
0	0.108	1.784\\
0	0.157	1.686\\
0	0.206	1.588\\
0	0.255	1.49\\
0	0.304	1.392\\
0	0.353	1.294\\
0	0.402	1.196\\
0	0.451	1.098\\
0	0.5	1.00000000160102\\
0	0.549	1.098\\
0	0.598	1.196\\
0	0.647	1.294\\
0	0.696	1.392\\
0	0.745	1.49\\
0	0.794	1.588\\
0	0.843	1.686\\
0	0.892	1.784\\
0	0.941	1.882\\
0	0.99	1.98\\
0.3	0.01	2.28\\
0.3	0.059	2.182\\
0.3	0.108	2.084\\
0.3	0.157	1.986\\
0.3	0.206	1.888\\
0.3	0.255	1.79\\
0.3	0.304	1.692\\
0.3	0.353	1.594\\
0.3	0.402	1.496\\
0.3	0.451	1.398\\
0.3	0.5	1.3\\
0.3	0.549	1.202\\
0.3	0.598	1.104\\
0.3	0.647	1.006\\
0.3	0.696	1.092\\
0.3	0.745	1.19\\
0.3	0.794	1.288\\
0.3	0.843	1.386\\
0.3	0.892	1.484\\
0.3	0.941	1.582\\
0.3	0.99	1.68\\
0.6	0.01	2.58\\
0.6	0.059	2.482\\
0.6	0.108	2.384\\
0.6	0.157	2.286\\
0.6	0.206	2.188\\
0.6	0.255	2.09\\
0.6	0.304	1.992\\
0.6	0.353	1.894\\
0.6	0.402	1.796\\
0.6	0.451	1.698\\
0.6	0.5	1.6\\
0.6	0.549	1.502\\
0.6	0.598	1.404\\
0.6	0.647	1.306\\
0.6	0.696	1.208\\
0.6	0.745	1.11\\
0.6	0.794	1.012\\
0.6	0.843	1.086\\
0.6	0.892	1.184\\
0.6	0.941	1.282\\
0.6	0.99	1.38\\
0.9	0.01	2.88\\
0.9	0.059	2.782\\
0.9	0.108	2.684\\
0.9	0.157	2.586\\
0.9	0.206	2.488\\
0.9	0.255	2.39\\
0.9	0.304	2.292\\
0.9	0.353	2.194\\
0.9	0.402	2.096\\
0.9	0.451	1.998\\
0.9	0.5	1.9\\
0.9	0.549	1.802\\
0.9	0.598	1.704\\
0.9	0.647	1.606\\
0.9	0.696	1.508\\
0.9	0.745	1.41\\
0.9	0.794	1.312\\
0.9	0.843	1.214\\
0.9	0.892	1.116\\
0.9	0.941	1.018\\
0.9	0.99	1.08\\
1.2	0.01	3.18\\
1.2	0.059	3.082\\
1.2	0.108	2.984\\
1.2	0.157	2.886\\
1.2	0.206	2.788\\
1.2	0.255	2.69\\
1.2	0.304	2.592\\
1.2	0.353	2.494\\
1.2	0.402	2.396\\
1.2	0.451	2.298\\
1.2	0.5	2.2\\
1.2	0.549	2.102\\
1.2	0.598	2.004\\
1.2	0.647	1.906\\
1.2	0.696	1.808\\
1.2	0.745	1.71\\
1.2	0.794	1.612\\
1.2	0.843	1.514\\
1.2	0.892	1.416\\
1.2	0.941	1.318\\
1.2	0.99	1.22\\
1.5	0.01	3.48\\
1.5	0.059	3.382\\
1.5	0.108	3.284\\
1.5	0.157	3.186\\
1.5	0.206	3.088\\
1.5	0.255	2.99\\
1.5	0.304	2.892\\
1.5	0.353	2.794\\
1.5	0.402	2.696\\
1.5	0.451	2.598\\
1.5	0.5	2.5\\
1.5	0.549	2.402\\
1.5	0.598	2.304\\
1.5	0.647	2.206\\
1.5	0.696	2.108\\
1.5	0.745	2.01\\
1.5	0.794	1.912\\
1.5	0.843	1.814\\
1.5	0.892	1.716\\
1.5	0.941	1.618\\
1.5	0.99	1.52\\
1.8	0.01	3.78\\
1.8	0.059	3.682\\
1.8	0.108	3.584\\
1.8	0.157	3.486\\
1.8	0.206	3.388\\
1.8	0.255	3.29\\
1.8	0.304	3.192\\
1.8	0.353	3.094\\
1.8	0.402	2.996\\
1.8	0.451	2.898\\
1.8	0.5	2.8\\
1.8	0.549	2.702\\
1.8	0.598	2.604\\
1.8	0.647	2.506\\
1.8	0.696	2.408\\
1.8	0.745	2.31\\
1.8	0.794	2.212\\
1.8	0.843	2.114\\
1.8	0.892	2.016\\
1.8	0.941	1.918\\
1.8	0.99	1.82\\
2.1	0.01	4.08\\
2.1	0.059	3.982\\
2.1	0.108	3.884\\
2.1	0.157	3.786\\
2.1	0.206	3.688\\
2.1	0.255	3.59\\
2.1	0.304	3.492\\
2.1	0.353	3.394\\
2.1	0.402	3.296\\
2.1	0.451	3.198\\
2.1	0.5	3.1\\
2.1	0.549	3.002\\
2.1	0.598	2.904\\
2.1	0.647	2.806\\
2.1	0.696	2.708\\
2.1	0.745	2.61\\
2.1	0.794	2.512\\
2.1	0.843	2.414\\
2.1	0.892	2.316\\
2.1	0.941	2.218\\
2.1	0.99	2.12\\
2.4	0.01	4.38\\
2.4	0.059	4.282\\
2.4	0.108	4.184\\
2.4	0.157	4.086\\
2.4	0.206	3.988\\
2.4	0.255	3.89\\
2.4	0.304	3.792\\
2.4	0.353	3.694\\
2.4	0.402	3.596\\
2.4	0.451	3.498\\
2.4	0.5	3.4\\
2.4	0.549	3.302\\
2.4	0.598	3.204\\
2.4	0.647	3.106\\
2.4	0.696	3.008\\
2.4	0.745	2.91\\
2.4	0.794	2.812\\
2.4	0.843	2.714\\
2.4	0.892	2.616\\
2.4	0.941	2.518\\
2.4	0.99	2.42\\
2.7	0.01	4.68\\
2.7	0.059	4.582\\
2.7	0.108	4.484\\
2.7	0.157	4.386\\
2.7	0.206	4.288\\
2.7	0.255	4.19\\
2.7	0.304	4.092\\
2.7	0.353	3.994\\
2.7	0.402	3.896\\
2.7	0.451	3.798\\
2.7	0.5	3.7\\
2.7	0.549	3.602\\
2.7	0.598	3.504\\
2.7	0.647	3.406\\
2.7	0.696	3.308\\
2.7	0.745	3.21\\
2.7	0.794	3.112\\
2.7	0.843	3.014\\
2.7	0.892	2.916\\
2.7	0.941	2.818\\
2.7	0.99	2.72\\
3	0.01	4.98\\
3	0.059	4.882\\
3	0.108	4.784\\
3	0.157	4.686\\
3	0.206	4.588\\
3	0.255	4.49\\
3	0.304	4.392\\
3	0.353	4.294\\
3	0.402	4.196\\
3	0.451	4.098\\
3	0.5	4\\
3	0.549	3.902\\
3	0.598	3.804\\
3	0.647	3.706\\
3	0.696	3.608\\
3	0.745	3.51\\
3	0.794	3.412\\
3	0.843	3.314\\
3	0.892	3.216\\
3	0.941	3.118\\
3	0.99	3.02\\
};

\addplot3[solid]
 table[row sep=crcr] {%
-3	0	3\\
-2.7	0	2.7\\
-2.4	0	2.4\\
-2.1	0	2.1\\
-1.8	0	1.8\\
-1.5	0	1.5\\
-1.2	0	1.2\\
-0.9	0	1.1\\
-0.6	0	1.4\\
-0.3	0	1.7\\
0	0	2\\
0.3	0	2.9\\
0.6	0	3.8\\
0.9	0	4.7\\
1.2	0	5.6\\
1.5	0	6.5\\
1.8	0	7.4\\
2.1	0	8.3\\
2.4	0	9.2\\
2.7	0	10.1\\
3	0	11\\
};

\addplot3[solid]
 table[row sep=crcr] {%
-3	1	11\\
-2.7	1	10.1\\
-2.4	1	9.2\\
-2.1	1	8.3\\
-1.8	1	7.4\\
-1.5	1	6.5\\
-1.2	1	5.6\\
-0.9	1	4.7\\
-0.6	1	3.8\\
-0.3	1	2.9\\
0	1	2\\
0.3	1	1.7\\
0.6	1	1.4\\
0.9	1	1.1\\
1.2	1	1.2\\
1.5	1	1.5\\
1.8	1	1.8\\
2.1	1	2.1\\
2.4	1	2.4\\
2.7	1	2.7\\
3	1	3\\
};
 \end{axis}
\end{tikzpicture}%
}\hfill
\subfigure[]{
\label{fig:L1-bounded-continuous}
\input{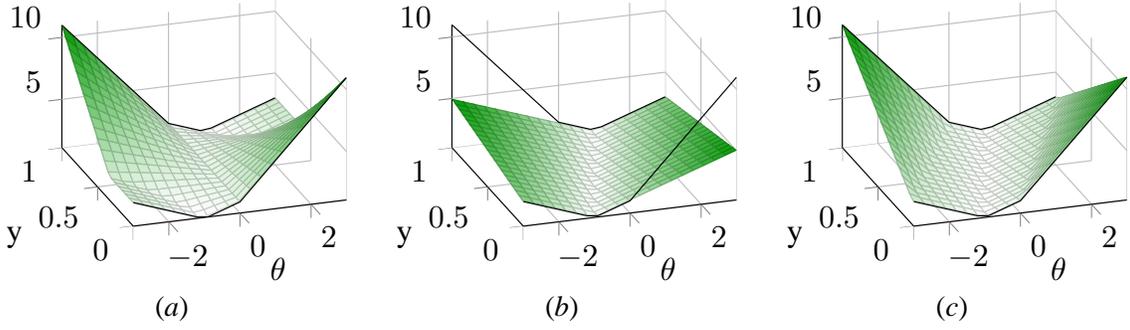}
}
\caption{Depicted above in (a) is the non-convex objective function $\varphi$ 
of the optimization problem
\eqref{eq:problem-main}
for $Y = \{0,1\}$, $\Theta = \mathbb{R}$, the Hinge loss 
(Tab.~\ref{table:loss-functions})
and $\omega(\cdot) = \|\cdot\|_1$.  
Its restriction $\phi$ to $\Theta \times \{0,1\}$ is depicted in black.
Depicted in (b) is the tightest convex extension of $\phi$ to $\Theta \times [0,1]$.
Depicted in (c) is the tightest convex extension of $\phi$ to $[-b,b] \times [0,1]$, for $b \in \mathbb{R}_0^+$.
It can be seen that, for bounded $\theta$, the tightest convex extension is continuous.}
\label{fig:L1-explanation}
\end{figure*} 
In this section, we characterize, for common loss and regularization functions, convex extensions $\phi'$ of $\phi$ that are not necessarily tight but can be computed efficiently in general.
The construction is in two steps:

Firstly, we consider an additive decomposition of $\phi$.
Specifically, we consider a convex function $c: \Theta \to \mathbb{R}_0^+$ and, for every $s \in S$, a function $d_s: \Theta \times \{0,1\} \to \mathbb{R}_0^+$ such that $d_s(\cdot, 0)$ and $d_s(\cdot, 1)$ are convex.
These functions are chosen such that, for all $(\theta, y) \in \Theta \times Y$:
\begin{align}
\phi(\theta, y) = c(\theta) + \frac{1}{|S|} \sum_{s \in S} d_s(\theta, y_s)
\enspace .
\label{eq:decomposition}
\end{align}
One example is given by $c := \omega$ and, for every $s \in S$, $d_s := C \, l_s$.
Another example is given by $c := 0$ and, for every $s \in S$, every $\theta \in \Theta$ and every $y_s \in \{0,1\}$: $d_s(\theta, y_s) := \omega(\theta) + C l_s(\theta, y_s)$.

Secondly, we characterize, for each $d_s$, its tightest convex extension $d^{**}_s: \Theta \times [0,1] \to \mathbb{R}$, by a specialization of 
Theorem~\ref{theorem:main}
stated as
Corollary~\ref{theorem:binary_envelope_theorem}.
A convex extension $\phi'$ of $\phi$ to $\Theta \times [0,1]^S$ is then given by 
\begin{align}
\phi': 
    \quad \Theta \times [0,1]^S \to \mathbb{R}_0^+:
    \quad (\theta, y) \mapsto c(\theta) + \frac{1}{|S|} \sum_{s \in S} d^{**}_s(\theta, y_s)
\enspace .
\end{align}

\begin{corollary} 
\label{theorem:binary_envelope_theorem} 
For every $d: \Theta \times \{0,1\} \to \mathbb{R}_0^+$ 
such that $d_0(\cdot) := d(\cdot, 0)$ and $d_1(\cdot) := d(\cdot, 1)$ are convex, 
the tightest convex extension $d^{**}: \Theta \times [0,1] \to \mathbb{R}_0^+$ of $d$ is such that 
for all $(\theta, y) \in \Theta \times Y$:
\begin{align}
\label{eq:binary_convex_envelope}
d^{**}(\theta, y) = \begin{cases} 
d(\theta, y) &\mbox{if } y \in \{0,1\} \\ 
\Psi(\theta, y) & \mbox{if } y \in (0,1)
\end{cases}
\end{align}
with
\begin{align}
\Psi(\theta, y) 
= 
\inf_{\theta^0, \theta^1 \in \Theta} \left\{
    (1-y) d_0(\theta^0)+y d_1(\theta^1)
    \ \middle| \ 
    (1-y) \theta^0 + y \theta^1 = \theta
\right\}
\enspace .
\label{eq:big_psi}
\end{align}
\end{corollary}

The solutions of the optimization problem \eqref{eq:big_psi} are characterized in Lemma~\ref{lemma:subgradient-1}.
Subgradients are given in 
Lemma~\ref{corollary:subgradient}.
\begin{lemma} 
\label{lemma:subgradient-1}
For every solution $(\hat\theta^0, \hat\theta^1) \in \Theta^2$ of \eqref{eq:big_psi}, 
there exists a $G \neq \emptyset$ such that, for $\xi_{\theta, y}: \Theta \to \Theta: t \mapsto (\theta-(1-y) t)/y$:
\begin{align}
(\delta d_0)(\hat\theta^0) \ \bigcap\ \ (\delta d_1)(\xi_{\theta, y}(\hat\theta^0)) & = G
    \label{eq:minimizer_convex_envelope}\\
(\delta d_0)(\xi_{\theta, 1-y}(\hat\theta^1)) \ \bigcap\ \ (\delta d_1)(\hat\theta^1) & = G
    \label{eq:minimizer_convex_envelope_equivalent}
\enspace .
\end{align}
\end{lemma}
\begin{lemma}
\label{corollary:subgradient}
If a solution $(\hat\theta^0, \hat\theta^1) \in \Theta^2$ of \eqref{eq:big_psi} exists,
then 
$\tbinom{v}{w} \in (\delta d^{**}) (\theta, y)$
iff
$v \in G$ with $G$ defined equivalently in 
\eqref{eq:minimizer_convex_envelope} and \eqref{eq:minimizer_convex_envelope_equivalent}
and $w = v^T\hat\theta^0 - v^T \hat\theta^1 + d_1(\hat\theta^1) - d_0(\hat\theta^0)$.
Otherwise, $y \in \{0,1\}$ and:
\begin{align}
y = 0 
    \quad \Rightarrow \quad
    \forall v \in (\delta d(\cdot, 0))(\theta): \ 
        & \tbinom{v}{\infty} \in (\delta d^{**}) (\theta, y) \\
y = 1
    \quad \Rightarrow \quad
    \forall v \in (\delta d(\cdot, 1))(\theta): \ 
        & \tbinom{v}{-\infty} \in (\delta d^{**}) (\theta, y)
\enspace .
\end{align}
\end{lemma}

\begin{figure*}
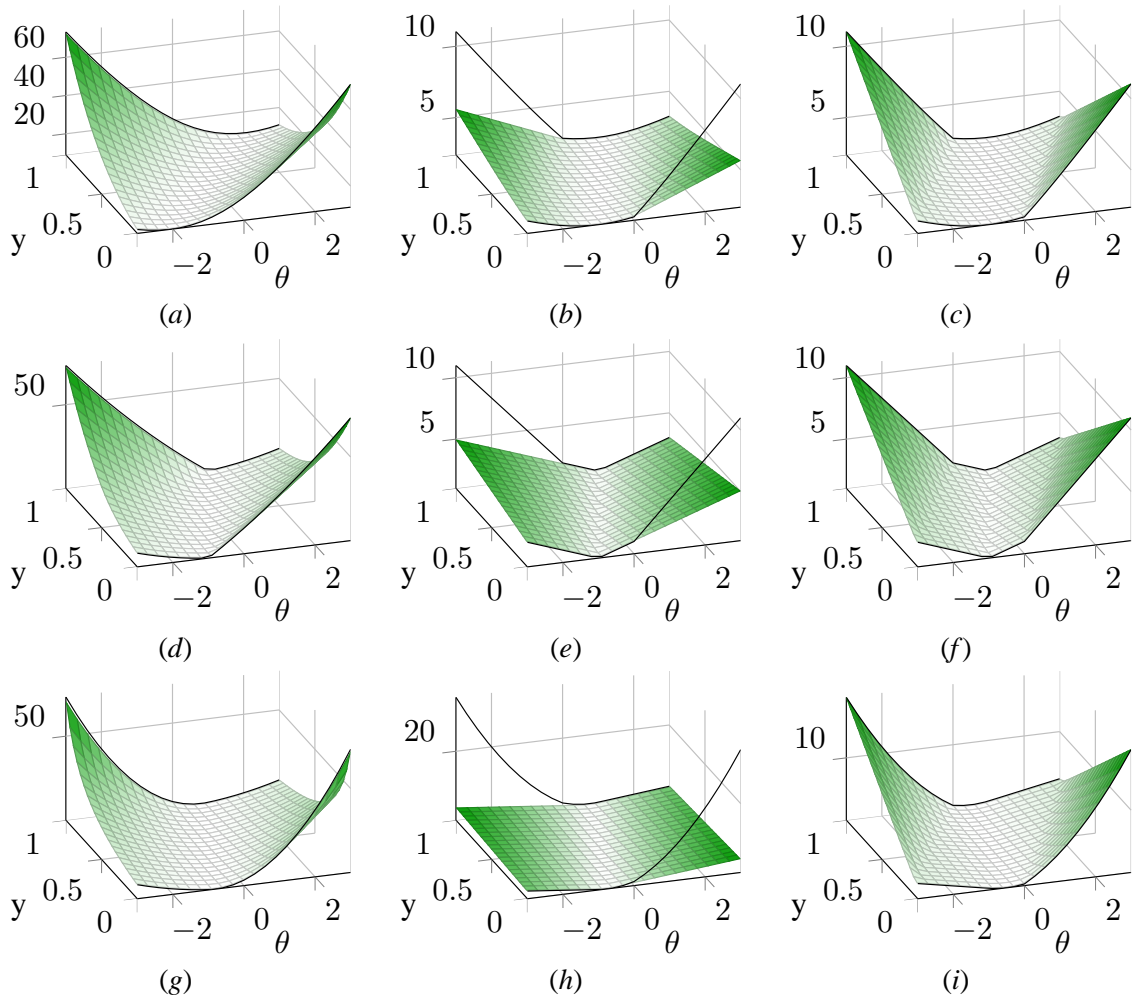

\subfigure[]{%
\label{fig:Envelope_Logistic_L2-plot}
\input{figures-2/Envelope_Logistic_L2_3d.tex}
}\hfill
\subfigure[]{%
\label{fig:Envelope_Logistic_L1-plot}
\input{figures-2/Envelope_Logistic_L1_3d.tex}
}\hfill
\subfigure[]{%
\label{fig:Envelope_Logistic_L1_bounded-plot}
\input{figures-2/Envelope_Logistic_L1_BND_3d.tex}
}\\
\subfigure[]{%
\label{fig:Envelope_H1_L2-plot}
\input{figures-2/Envelope_H1_L2_3d.tex}
}\hfill
\subfigure[]{%
\label{fig:Envelope_H1_L1-plot}
%
%
\begin{tikzpicture}

\begin{axis}[%
width=\smallplotwidth,
every outer x axis line/.append style={black},
every x tick label/.append style={font=\color{black}},
xmin=-3,
xmax=3,
tick align=outside,
xlabel={$\theta$},
    xlabel style={yshift=3ex},
xmajorgrids,
every outer y axis line/.append style={black},
every y tick label/.append style={font=\color{black}},
ymin=0,
ymax=1,
ylabel={y},
    ylabel style={yshift=3ex},
ymajorgrids,
every outer z axis line/.append style={black},
every z tick label/.append style={font=\color{black}},
zmin=1.00000000160102,
zmax=11,
zmajorgrids,
view={-18.5}{34},
axis x line*=bottom,
axis y line*=left,
axis z line*=left,
colormap={mycolormap}{color(0)=(white) color(1)=(mycolor1)}
]

\addplot3[%
    surf,
    shader=faceted,
    mesh/rows=21
]
table[row sep=crcr,header=false] {%
-3	0.01	3.02\\
-3	0.059	3.118\\
-3	0.108	3.216\\
-3	0.157	3.314\\
-3	0.206	3.412\\
-3	0.255	3.51\\
-3	0.304	3.608\\
-3	0.353	3.706\\
-3	0.402	3.804\\
-3	0.451	3.902\\
-3	0.5	4\\
-3	0.549	4.098\\
-3	0.598	4.196\\
-3	0.647	4.294\\
-3	0.696	4.392\\
-3	0.745	4.49\\
-3	0.794	4.588\\
-3	0.843	4.686\\
-3	0.892	4.784\\
-3	0.941	4.882\\
-3	0.99	4.98\\
-2.7	0.01	2.72\\
-2.7	0.059	2.818\\
-2.7	0.108	2.916\\
-2.7	0.157	3.014\\
-2.7	0.206	3.112\\
-2.7	0.255	3.21\\
-2.7	0.304	3.308\\
-2.7	0.353	3.406\\
-2.7	0.402	3.504\\
-2.7	0.451	3.602\\
-2.7	0.5	3.7\\
-2.7	0.549	3.798\\
-2.7	0.598	3.896\\
-2.7	0.647	3.994\\
-2.7	0.696	4.092\\
-2.7	0.745	4.19\\
-2.7	0.794	4.288\\
-2.7	0.843	4.386\\
-2.7	0.892	4.484\\
-2.7	0.941	4.582\\
-2.7	0.99	4.68\\
-2.4	0.01	2.42\\
-2.4	0.059	2.518\\
-2.4	0.108	2.616\\
-2.4	0.157	2.714\\
-2.4	0.206	2.812\\
-2.4	0.255	2.91\\
-2.4	0.304	3.008\\
-2.4	0.353	3.106\\
-2.4	0.402	3.204\\
-2.4	0.451	3.302\\
-2.4	0.5	3.4\\
-2.4	0.549	3.498\\
-2.4	0.598	3.596\\
-2.4	0.647	3.694\\
-2.4	0.696	3.792\\
-2.4	0.745	3.89\\
-2.4	0.794	3.988\\
-2.4	0.843	4.086\\
-2.4	0.892	4.184\\
-2.4	0.941	4.282\\
-2.4	0.99	4.38\\
-2.1	0.01	2.12\\
-2.1	0.059	2.218\\
-2.1	0.108	2.316\\
-2.1	0.157	2.414\\
-2.1	0.206	2.512\\
-2.1	0.255	2.61\\
-2.1	0.304	2.708\\
-2.1	0.353	2.806\\
-2.1	0.402	2.904\\
-2.1	0.451	3.002\\
-2.1	0.5	3.1\\
-2.1	0.549	3.198\\
-2.1	0.598	3.296\\
-2.1	0.647	3.394\\
-2.1	0.696	3.492\\
-2.1	0.745	3.59\\
-2.1	0.794	3.688\\
-2.1	0.843	3.786\\
-2.1	0.892	3.884\\
-2.1	0.941	3.982\\
-2.1	0.99	4.08\\
-1.8	0.01	1.82\\
-1.8	0.059	1.918\\
-1.8	0.108	2.016\\
-1.8	0.157	2.114\\
-1.8	0.206	2.212\\
-1.8	0.255	2.31\\
-1.8	0.304	2.408\\
-1.8	0.353	2.506\\
-1.8	0.402	2.604\\
-1.8	0.451	2.702\\
-1.8	0.5	2.8\\
-1.8	0.549	2.898\\
-1.8	0.598	2.996\\
-1.8	0.647	3.094\\
-1.8	0.696	3.192\\
-1.8	0.745	3.29\\
-1.8	0.794	3.388\\
-1.8	0.843	3.486\\
-1.8	0.892	3.584\\
-1.8	0.941	3.682\\
-1.8	0.99	3.78\\
-1.5	0.01	1.52\\
-1.5	0.059	1.618\\
-1.5	0.108	1.716\\
-1.5	0.157	1.814\\
-1.5	0.206	1.912\\
-1.5	0.255	2.01\\
-1.5	0.304	2.108\\
-1.5	0.353	2.206\\
-1.5	0.402	2.304\\
-1.5	0.451	2.402\\
-1.5	0.5	2.5\\
-1.5	0.549	2.598\\
-1.5	0.598	2.696\\
-1.5	0.647	2.794\\
-1.5	0.696	2.892\\
-1.5	0.745	2.99\\
-1.5	0.794	3.088\\
-1.5	0.843	3.186\\
-1.5	0.892	3.284\\
-1.5	0.941	3.382\\
-1.5	0.99	3.48\\
-1.2	0.01	1.22\\
-1.2	0.059	1.318\\
-1.2	0.108	1.416\\
-1.2	0.157	1.514\\
-1.2	0.206	1.612\\
-1.2	0.255	1.71\\
-1.2	0.304	1.808\\
-1.2	0.353	1.906\\
-1.2	0.402	2.004\\
-1.2	0.451	2.102\\
-1.2	0.5	2.2\\
-1.2	0.549	2.298\\
-1.2	0.598	2.396\\
-1.2	0.647	2.494\\
-1.2	0.696	2.592\\
-1.2	0.745	2.69\\
-1.2	0.794	2.788\\
-1.2	0.843	2.886\\
-1.2	0.892	2.984\\
-1.2	0.941	3.082\\
-1.2	0.99	3.18\\
-0.9	0.01	1.08\\
-0.9	0.059	1.018\\
-0.9	0.108	1.116\\
-0.9	0.157	1.214\\
-0.9	0.206	1.312\\
-0.9	0.255	1.41\\
-0.9	0.304	1.508\\
-0.9	0.353	1.606\\
-0.9	0.402	1.704\\
-0.9	0.451	1.802\\
-0.9	0.5	1.9\\
-0.9	0.549	1.998\\
-0.9	0.598	2.096\\
-0.9	0.647	2.194\\
-0.9	0.696	2.292\\
-0.9	0.745	2.39\\
-0.9	0.794	2.488\\
-0.9	0.843	2.586\\
-0.9	0.892	2.684\\
-0.9	0.941	2.782\\
-0.9	0.99	2.88\\
-0.6	0.01	1.38\\
-0.6	0.059	1.282\\
-0.6	0.108	1.184\\
-0.6	0.157	1.086\\
-0.6	0.206	1.012\\
-0.6	0.255	1.11\\
-0.6	0.304	1.208\\
-0.6	0.353	1.306\\
-0.6	0.402	1.404\\
-0.6	0.451	1.502\\
-0.6	0.5	1.6\\
-0.6	0.549	1.698\\
-0.6	0.598	1.796\\
-0.6	0.647	1.894\\
-0.6	0.696	1.992\\
-0.6	0.745	2.09\\
-0.6	0.794	2.188\\
-0.6	0.843	2.286\\
-0.6	0.892	2.384\\
-0.6	0.941	2.482\\
-0.6	0.99	2.58\\
-0.3	0.01	1.68\\
-0.3	0.059	1.582\\
-0.3	0.108	1.484\\
-0.3	0.157	1.386\\
-0.3	0.206	1.288\\
-0.3	0.255	1.19\\
-0.3	0.304	1.092\\
-0.3	0.353	1.006\\
-0.3	0.402	1.104\\
-0.3	0.451	1.202\\
-0.3	0.5	1.3\\
-0.3	0.549	1.398\\
-0.3	0.598	1.496\\
-0.3	0.647	1.594\\
-0.3	0.696	1.692\\
-0.3	0.745	1.79\\
-0.3	0.794	1.888\\
-0.3	0.843	1.986\\
-0.3	0.892	2.084\\
-0.3	0.941	2.182\\
-0.3	0.99	2.28\\
0	0.01	1.98\\
0	0.059	1.882\\
0	0.108	1.784\\
0	0.157	1.686\\
0	0.206	1.588\\
0	0.255	1.49\\
0	0.304	1.392\\
0	0.353	1.294\\
0	0.402	1.196\\
0	0.451	1.098\\
0	0.5	1.00000000160102\\
0	0.549	1.098\\
0	0.598	1.196\\
0	0.647	1.294\\
0	0.696	1.392\\
0	0.745	1.49\\
0	0.794	1.588\\
0	0.843	1.686\\
0	0.892	1.784\\
0	0.941	1.882\\
0	0.99	1.98\\
0.3	0.01	2.28\\
0.3	0.059	2.182\\
0.3	0.108	2.084\\
0.3	0.157	1.986\\
0.3	0.206	1.888\\
0.3	0.255	1.79\\
0.3	0.304	1.692\\
0.3	0.353	1.594\\
0.3	0.402	1.496\\
0.3	0.451	1.398\\
0.3	0.5	1.3\\
0.3	0.549	1.202\\
0.3	0.598	1.104\\
0.3	0.647	1.006\\
0.3	0.696	1.092\\
0.3	0.745	1.19\\
0.3	0.794	1.288\\
0.3	0.843	1.386\\
0.3	0.892	1.484\\
0.3	0.941	1.582\\
0.3	0.99	1.68\\
0.6	0.01	2.58\\
0.6	0.059	2.482\\
0.6	0.108	2.384\\
0.6	0.157	2.286\\
0.6	0.206	2.188\\
0.6	0.255	2.09\\
0.6	0.304	1.992\\
0.6	0.353	1.894\\
0.6	0.402	1.796\\
0.6	0.451	1.698\\
0.6	0.5	1.6\\
0.6	0.549	1.502\\
0.6	0.598	1.404\\
0.6	0.647	1.306\\
0.6	0.696	1.208\\
0.6	0.745	1.11\\
0.6	0.794	1.012\\
0.6	0.843	1.086\\
0.6	0.892	1.184\\
0.6	0.941	1.282\\
0.6	0.99	1.38\\
0.9	0.01	2.88\\
0.9	0.059	2.782\\
0.9	0.108	2.684\\
0.9	0.157	2.586\\
0.9	0.206	2.488\\
0.9	0.255	2.39\\
0.9	0.304	2.292\\
0.9	0.353	2.194\\
0.9	0.402	2.096\\
0.9	0.451	1.998\\
0.9	0.5	1.9\\
0.9	0.549	1.802\\
0.9	0.598	1.704\\
0.9	0.647	1.606\\
0.9	0.696	1.508\\
0.9	0.745	1.41\\
0.9	0.794	1.312\\
0.9	0.843	1.214\\
0.9	0.892	1.116\\
0.9	0.941	1.018\\
0.9	0.99	1.08\\
1.2	0.01	3.18\\
1.2	0.059	3.082\\
1.2	0.108	2.984\\
1.2	0.157	2.886\\
1.2	0.206	2.788\\
1.2	0.255	2.69\\
1.2	0.304	2.592\\
1.2	0.353	2.494\\
1.2	0.402	2.396\\
1.2	0.451	2.298\\
1.2	0.5	2.2\\
1.2	0.549	2.102\\
1.2	0.598	2.004\\
1.2	0.647	1.906\\
1.2	0.696	1.808\\
1.2	0.745	1.71\\
1.2	0.794	1.612\\
1.2	0.843	1.514\\
1.2	0.892	1.416\\
1.2	0.941	1.318\\
1.2	0.99	1.22\\
1.5	0.01	3.48\\
1.5	0.059	3.382\\
1.5	0.108	3.284\\
1.5	0.157	3.186\\
1.5	0.206	3.088\\
1.5	0.255	2.99\\
1.5	0.304	2.892\\
1.5	0.353	2.794\\
1.5	0.402	2.696\\
1.5	0.451	2.598\\
1.5	0.5	2.5\\
1.5	0.549	2.402\\
1.5	0.598	2.304\\
1.5	0.647	2.206\\
1.5	0.696	2.108\\
1.5	0.745	2.01\\
1.5	0.794	1.912\\
1.5	0.843	1.814\\
1.5	0.892	1.716\\
1.5	0.941	1.618\\
1.5	0.99	1.52\\
1.8	0.01	3.78\\
1.8	0.059	3.682\\
1.8	0.108	3.584\\
1.8	0.157	3.486\\
1.8	0.206	3.388\\
1.8	0.255	3.29\\
1.8	0.304	3.192\\
1.8	0.353	3.094\\
1.8	0.402	2.996\\
1.8	0.451	2.898\\
1.8	0.5	2.8\\
1.8	0.549	2.702\\
1.8	0.598	2.604\\
1.8	0.647	2.506\\
1.8	0.696	2.408\\
1.8	0.745	2.31\\
1.8	0.794	2.212\\
1.8	0.843	2.114\\
1.8	0.892	2.016\\
1.8	0.941	1.918\\
1.8	0.99	1.82\\
2.1	0.01	4.08\\
2.1	0.059	3.982\\
2.1	0.108	3.884\\
2.1	0.157	3.786\\
2.1	0.206	3.688\\
2.1	0.255	3.59\\
2.1	0.304	3.492\\
2.1	0.353	3.394\\
2.1	0.402	3.296\\
2.1	0.451	3.198\\
2.1	0.5	3.1\\
2.1	0.549	3.002\\
2.1	0.598	2.904\\
2.1	0.647	2.806\\
2.1	0.696	2.708\\
2.1	0.745	2.61\\
2.1	0.794	2.512\\
2.1	0.843	2.414\\
2.1	0.892	2.316\\
2.1	0.941	2.218\\
2.1	0.99	2.12\\
2.4	0.01	4.38\\
2.4	0.059	4.282\\
2.4	0.108	4.184\\
2.4	0.157	4.086\\
2.4	0.206	3.988\\
2.4	0.255	3.89\\
2.4	0.304	3.792\\
2.4	0.353	3.694\\
2.4	0.402	3.596\\
2.4	0.451	3.498\\
2.4	0.5	3.4\\
2.4	0.549	3.302\\
2.4	0.598	3.204\\
2.4	0.647	3.106\\
2.4	0.696	3.008\\
2.4	0.745	2.91\\
2.4	0.794	2.812\\
2.4	0.843	2.714\\
2.4	0.892	2.616\\
2.4	0.941	2.518\\
2.4	0.99	2.42\\
2.7	0.01	4.68\\
2.7	0.059	4.582\\
2.7	0.108	4.484\\
2.7	0.157	4.386\\
2.7	0.206	4.288\\
2.7	0.255	4.19\\
2.7	0.304	4.092\\
2.7	0.353	3.994\\
2.7	0.402	3.896\\
2.7	0.451	3.798\\
2.7	0.5	3.7\\
2.7	0.549	3.602\\
2.7	0.598	3.504\\
2.7	0.647	3.406\\
2.7	0.696	3.308\\
2.7	0.745	3.21\\
2.7	0.794	3.112\\
2.7	0.843	3.014\\
2.7	0.892	2.916\\
2.7	0.941	2.818\\
2.7	0.99	2.72\\
3	0.01	4.98\\
3	0.059	4.882\\
3	0.108	4.784\\
3	0.157	4.686\\
3	0.206	4.588\\
3	0.255	4.49\\
3	0.304	4.392\\
3	0.353	4.294\\
3	0.402	4.196\\
3	0.451	4.098\\
3	0.5	4\\
3	0.549	3.902\\
3	0.598	3.804\\
3	0.647	3.706\\
3	0.696	3.608\\
3	0.745	3.51\\
3	0.794	3.412\\
3	0.843	3.314\\
3	0.892	3.216\\
3	0.941	3.118\\
3	0.99	3.02\\
};
\addplot3 [solid]
 table[row sep=crcr] {%
-3	0	3\\
-2.7	0	2.7\\
-2.4	0	2.4\\
-2.1	0	2.1\\
-1.8	0	1.8\\
-1.5	0	1.5\\
-1.2	0	1.2\\
-0.9	0	1.1\\
-0.6	0	1.4\\
-0.3	0	1.7\\
0	0	2\\
0.3	0	2.9\\
0.6	0	3.8\\
0.9	0	4.7\\
1.2	0	5.6\\
1.5	0	6.5\\
1.8	0	7.4\\
2.1	0	8.3\\
2.4	0	9.2\\
2.7	0	10.1\\
3	0	11\\
};
 \addplot3 [solid]
 table[row sep=crcr] {%
-3	1	11\\
-2.7	1	10.1\\
-2.4	1	9.2\\
-2.1	1	8.3\\
-1.8	1	7.4\\
-1.5	1	6.5\\
-1.2	1	5.6\\
-0.9	1	4.7\\
-0.6	1	3.8\\
-0.3	1	2.9\\
0	1	2\\
0.3	1	1.7\\
0.6	1	1.4\\
0.9	1	1.1\\
1.2	1	1.2\\
1.5	1	1.5\\
1.8	1	1.8\\
2.1	1	2.1\\
2.4	1	2.4\\
2.7	1	2.7\\
3	1	3\\
};
 \end{axis}
\end{tikzpicture}%
}\hfill
\subfigure[]{%
\label{fig:Envelope_H1_L1_bounded-plot}
\input{figures-2/Envelope_H1_L1_BND_3d.tex}
}\\
\subfigure[]{%
\label{fig:Envelope_H2_L2-plot}
\input{figures-2/Envelope_H2_L2_3d.tex}
}\hfill
\subfigure[]{%
\label{fig:Envelope_H2_L1-plot}
\input{figures-2/Envelope_H2_L1_3d.tex}
}\hfill
\subfigure[]{%
\label{fig:Envelope_H2_L1_bounded-plot}
\input{figures-2/Envelope_H2_L1_BND_3d.tex}
}
\caption{Depicted above are tightest convex extensions of 
the logistic loss (first row), 
the Hinge loss (second row) and 
the squared Hinge loss (third row),
in conjunction with 
L2 regularization (first column),
L1 regularization (second column) and
L1 regularization with $\theta$ constrained to $[-3.1, 3.1]$ (third column).
Values for $y \in \{0,1\}$ are depicted in black.
Note that the convex extensions of L1-regularized loss functions for unbounded $\theta$ are discontinuous. 
Parameters for these examples are $x = 1$ and
\ref{fig:Envelope_Logistic_L2-plot}: $ C = 16 $, 
\ref{fig:Envelope_Logistic_L1-plot}: $ C = 5 $,
\ref{fig:Envelope_Logistic_L1_bounded-plot}: $ C = 5 $,
\ref{fig:Envelope_H1_L2-plot}: $ C = 16 $, 
\ref{fig:Envelope_H1_L1-plot}: $ C = 5 $, 
\ref{fig:Envelope_H1_L1_bounded-plot}: $ C = 5 $,
\ref{fig:Envelope_H2_L2-plot}: $ C = 4 $, 
\ref{fig:Envelope_H2_L1-plot}: $ C = 4 $.
\ref{fig:Envelope_H2_L1_bounded-plot}: $ C = 4 $.}
\label{fig:all_envelopes}
\end{figure*}

We proceed as follows:
In Section~\ref{section:simple-examples},
we consider two decompositions of the class
\eqref{eq:decomposition}
for which the tightest convex extension of any $d_s$ is easy to characterize but the resulting convex extension of $\phi$ is rather loose.
In Sections~\ref{section:l2} and \ref{section:l1},
we consider decompositions of the class
\eqref{eq:decomposition}
with $c = 0$.
These yield the tightest convex extension of $\phi$ of any decomposition of the class
\eqref{eq:decomposition}.
For any combination of the logistic loss, the Hinge loss and the squadred Hinge loss 
(Tab.~\ref{table:loss-functions})
with either L1 regularization $\omega(\cdot) = \|\cdot\|_1$ or L2 regularization  $\omega(\cdot) = \|\cdot\|_2^2$, we characterize the convex extension explicitly and, in some cases, in closed form.
Examples of all combinations are depicted in 
Fig.~\ref{fig:all_envelopes}.

\subsection{Instructive Examples}
\label{section:simple-examples}

The first convex extension we characterize is for the decomposition 
\eqref{eq:decomposition} 
with $c := \omega$ and $\forall s \in S: d_s := C l_s$.

\begin{corollary}
\label{corollary:trivial-convex-envelope}
For any $d: \Theta \times \{0,1\} \to \mathbb{R}_0^+$ 
such that $d_0(\cdot) := d(\cdot, 0)$ and $d_1(\cdot) := d(\cdot, 1)$ are convex
and for which $d_0(-r) \to 0$ and $d_1(r) \to 0$ as $r \to \infty$,
the convex extension $d^{**}: \Theta \times [0,1] \to \mathbb{R}$ of $d$ has the form 
\begin{align}
d'(\theta, y) = \begin{cases} 
d(\theta, y) &\mbox{if } y \in \{0,1\} \\ 
0 & \mbox{if } y \in (0,1)
\end{cases} 
\enspace .
\end{align}
\end{corollary}

The second convex extension we characterize is for the logistic loss
$l_s(\theta, y) = - \langle x, \theta \rangle y + \log(1+e^{\langle x,\theta \rangle})$,
for $\omega(\cdot) = \|\cdot\|_2^2$
and for the decomposition 
\eqref{eq:decomposition} 
such that, for every $s \in S$,
$d_s(\theta, y) := \|\theta\|_2^2 - C \langle x, \theta \rangle y$.

\begin{corollary}
\label{corollary:instructive-2}
The tightest convex extension 
$d^{**}: \Theta \times [0,1] \to \mathbb{R}_0^+$
of
$d: \Theta \times \{0,1\} \to \mathbb{R}_0^+$
with
$d(\theta, y) = \| \theta \|_2^2 - C \langle x, \theta \rangle y$
has the form
$d^{**}(\theta, y) = ||\theta -y \frac{C}{2} x ||_2^2 -y \|\frac{C}{2} x\|_2^2$.
\end{corollary}

\subsection{L2 Regularization}
\label{section:l2}
We now consider $\omega(\cdot) = \frac{1}{2} \|\cdot\|_2^2$
and decompositions
\eqref{eq:decomposition}
such that, for every $s \in S$: $d_s(\theta, y) = \omega(\theta) + C\,l_s(\theta, y)$.

\begin{corollary} 
\label{corollary:L2-general}
The tightest convex extension $d^{**}: \Theta \times [0,1] \to \mathbb{R}_0^+ $ of $ d : \Theta \times \{0,1\} \to \mathbb{R}_0^+ $ with $d(\theta,y) = \tfrac{1}{2} ||\theta||_2^2 + C l ( \langle x , \theta \rangle , y )$ is given by 
\eqref{eq:binary_convex_envelope} and \eqref{eq:big_psi}.
Moreover, the solution of \eqref{eq:big_psi} is given 
by $\hat\theta^0 = \theta - y C x z$ with
\begin{align}
z \  \in \ 
(\delta l_0)(x^T(\theta-y C x z))
- (\delta l_1)(x^T(\theta+(1-y) C x z)) 
\  \subseteq \  \mathbb{R}
\enspace .
\label{eq:L2-general}
\end{align}
\end{corollary}

Although this characterization is not a closed form, values of $d^{**}$ can be computed efficiently using the bisection method. For specific loss functions, closed forms are derived below.
\begin{corollary}
\label{corollary:H1L2}
The tightest convex extension
$d^{**}: \Theta \times [0,1] \to \mathbb{R}_0^+ $ 
of 
$d: \Theta \times \{0,1\} \to \mathbb{R}_0^+ $ 
with $d(\theta,y) = \tfrac{1}{2} ||\theta||_2^2 + C l ( \langle x , \theta \rangle , y )$
and $l : \mathbb{R} \times \{0,1\} \to \mathbb{R}_0^+$ of the form
\begin{align}
l(r, y) = (C_0(1-y) + C_1 y) \max \{0, 1 - (2y-1) r\}
\end{align}
where $C_0, C_1 \in \mathbb{R}_+$ define weights on the two values of $y$,
is given by 
Corollary~\ref{corollary:L2-general}.
Moreover, every $z$ satisfying
\eqref{eq:L2-general} 
holds
\begin{align}
z \in \left\{
0, \ 
C_0, \ 
C_1, \ 
C_0 + C_1, \ 
\frac{1+x^T \theta}{ y C x^T x}, \ 
\frac{1-x^T \theta}{(1-y) C x^T x}
\right\}
\enspace .
\end{align}
\end{corollary}
\begin{corollary}
\label{corollary:H2L2}
The tightest convex extension
$d^{**}: \Theta \times [0,1] \to \mathbb{R}_0^+ $ 
of 
$d : \Theta \times \{0,1\} \to \mathbb{R}_0^+$ 
with
$d(\theta, y) = \tfrac{1}{2} ||\theta||_2^2 + C l ( \langle x , \theta \rangle , y )$
and $l : \mathbb{R} \times \{0,1\} \to \mathbb{R}_0^+$ of the form
\begin{align}
l(r, y) = \frac{C_0(1-y) + C_1 y}{2} (\max \{0, 1 - (2y-1) r\})^2
\end{align}
where $C_0, C_1 \in \mathbb{R}_+$ define weights on the two values of $y$,
is given by 
Corollary~\ref{corollary:L2-general}.
Moreover, every $z$ satisfying
\eqref{eq:L2-general} 
holds
\begin{align}
z \in \left\{
0, \ 
\frac{C_0 + C_0 x^T \theta}{1 +  C_0 y C x^T x}, \ 
\frac{C_1-C_1 x^T \theta }{ 1+  C_1 (1-y) C x^T x}, \ 
\frac{C_0+C_1+(C_0-C_1) x^T \theta}{1+(C_0  y + C_1 (1-y)) C x^T x}
\right\}
\enspace .
\end{align}
\end{corollary}

\subsection{L1 regularization}
\label{section:l1}
We now consider $\omega(\cdot) = \|\cdot\|_1$
and decompositions
\eqref{eq:decomposition}
such that, for every $s \in S$:
$d_s(\theta, y) = \omega(\theta) + C\,l_s(\theta, y)$. 
We focus on a special case where $\theta$ is bounded. 
This is necessary in order for the convex extension to be continuous;
see Fig.~\ref{fig:L1-explanation}.

\begin{corollary} 
\label{corollary:L1-general}
For $b, t \in (\mathbb{R} \cup \{-\infty, \infty\})^m$ 
and $\Theta = \{\theta \in \mathbb{R}^m : b \leq \theta \leq t \}$,
the tightest convex extension $d^{**}: \Theta \times [0,1] \to \mathbb{R}_0^+$ 
of $d: \Theta \times \{0,1\} \to \mathbb{R}_0^+$ 
with $d(\theta,y) = ||\theta||_1 + C l (\langle x , \theta \rangle , y)$ 
is given by \eqref{eq:binary_convex_envelope} and \eqref{eq:big_psi}.
Moreover, the solution of \eqref{eq:big_psi} is given by
\begin{align}
\hat\theta^0 = \begin{cases}
    \theta' 
        & \mbox{if } \exists r \in  a(x^T \theta'): ||r C x ||_\infty \leq 2 \\
    \textnormal{aux}(\theta', x, a, b, t)
        & \mbox{otherwise}
\end{cases}
\end{align}
with $a: \mathbb{R} \to 2^\mathbb{R}: p \mapsto (\delta l_0)( p) - (\delta l_1)( (x^T \theta-(1-y) p)/y)$
and
\begin{align}
\label{eq:L1-projection-th}
\forall i \in \{1,\ldots,m\}: 
\quad
\theta'_i = \begin{cases} 
    \min\limits_{r \in [b'_i, t'_i]} |r| &\mbox{if } \theta_i > 0 \Leftrightarrow x_i > 0\\ 
    \min\limits_{r \in [b'_i, t'_i]} \left| \dfrac{\theta_i}{(1-y)} - r \right| &\mbox{otherwise}
\end{cases}
\enspace .
\end{align}
\end{corollary}
The function ``aux'' is defined below in terms of Alg.~\ref{algorithm:aux}.
At its core, this algorithms solves, for fixed $k \in [m]$ and fixed $V \in \Theta$,
the equation \eqref{eq:L1-general} for the unknown $z \in \mathbb{R}$.
For specific loss functions, closed forms are derived 
in Corollaries~\ref{corollary:H1L1} and \ref{corollary:H2L1}.
\begin{align}
\left| a\left(x^T V - z x_k\right) C x_k \right| = 2 
\label{eq:L1-general}
\end{align}
\begin{corollary}
\label{corollary:H1L1}
The tightest convex extension $d^{**}: \Theta \times [0,1] \to \mathbb{R}_0^+$ 
of $d: \Theta \times \{0,1\} \to \mathbb{R}_0^+$ 
with $d(\theta,y) = ||\theta||_1 + C l (\langle x , \theta \rangle , y)$
and $l: \mathbb{R} \times \{0,1\} \to \mathbb{R}_0^+$ of the form
\begin{align}
l(r, y) = (C_0(1-y) + C_1 y) \max \{0, 1 - (2y-1) r \}
\end{align}
where $C_0, C_1 \in \mathbb{R}_+$ define weights on the two values of $y$,
is given by Corollary~\ref{corollary:L1-general}.
Moreover, every solution $z \in \mathbb{R}$ of \eqref{eq:L1-general} holds
\begin{align}
z \in \left\{
\frac{1+x^T V}{x_k^2}, \ 
\frac{y+x^T (\theta - (1-y) V)}{(1-y) x_k^2}
\right \}
\enspace .
\end{align}
\end{corollary}
\begin{corollary}
\label{corollary:H2L1}
The tightest convex extension $d^{**}: \Theta \times [0,1] \to \mathbb{R}_0^+$ 
of $d: \Theta \times \{0,1\} \to \mathbb{R}_0^+$ 
with $d(\theta,y) = ||\theta||_1 + C l (\langle x, \theta \rangle , y)$
and $l : \mathbb{R} \times \{0,1\} \to \mathbb{R}_0^+$ of the form
\begin{align}
l(r, y) = \frac{C_0(1-y) + C_1 y}{2} (\max \{0, 1 - (2y-1) r \})^2
\end{align}
where $C_0, C_1 \in \mathbb{R}_+$ define weights on the two values of $y$,
is given by Corollary~\ref{corollary:L1-general}.
Moreover, every solution $z \in \mathbb{R}$ of \eqref{eq:L1-general} holds
\eqref{eq:z} with $v_0 := x^T V$ and $v_1 := - x^T \xi_{\theta,y}(V)$.
\begin{align}
z \in \bigg\{ &
	\frac{1+v_0 - 2 (C |x_k| C_0)^{-1}}{x_k^2}, \  
	\frac{1+v_1 - 2 (C |x_k| C_1)^{-1}}{x_k^2}, \nonumber\\
	&
	\frac{ y(C_0 (1 + v_0) + C_1 (1+v_1) - 2 (C |x_k|)^{-1}) }{(y C_0 + (1-y) C_1) x_k^2}
	\bigg\}
\label{eq:z}
\end{align}
\end{corollary}

\begin{algorithm}
\KwIn{$ \theta',x \in \mathbb{R}^n, a: \mathbb{R} \to 2^\mathbb{R}, C \in \mathbb{R}_0^+, b,t \in (\mathbb{R} \cup \{-\infty, \infty\})^n $}
\KwOut{$V \in \Theta$ }
$V := \theta' $ \;

$\forall i \in [n]: b'_i := \max\{b_i, (\xi_{\theta, 1-y}(t))_i\}$ \;

$\forall i \in [n]: t'_i := \min\{t_i, (\xi_{\theta, 1-y}(b))_i\}$ \;

$I \in \mathbb{N}^n $ such that $ |x_{I_1}| \geq |x_{I_2}| \geq ... \geq |x_{I_n}|  $ \;

\For{$j = 1 ... n $}{
    \If{$ \exists r \in a(x^T V): || r C x ||_\infty \leq 2 $}{
   		\Return $ V $\;
    } 
    $ i := I_j $\;

    \eIf{$ x_i > 0 $}{
        $ V_i := b'_i $\;
    }{
        $ V_i := t'_i $\;
    }
    \If{$ \exists r \in a(x^T V): || r C x ||_\infty \leq 2 $}{
        $V_i := V_i - z x_k $ with $ z \in \mathbb{R} $ such that $|a(x^T V - z x_k ) C x_k| = 2$\;

        \Return $ V $\;
    }
}
\Return $ V $\;
\caption{Computation of the function ``aux''}
\label{algorithm:aux}
\end{algorithm}

\section{Conclusion}
\label{section:conclusion}
We have characterized convex extensions, including the tightest convex extension,
of functions $f: \Theta \times Y \to \mathbb{R}_0^+$
with $\Theta \subseteq \mathbb{R}^m$ convex,
$Y \subseteq \{0,1\}^n$
and $f(\cdot, y)$ convex for every $y \in \{0,1\}^n$.
This has allowed us to state regularized empirical risk minimization with constrained labels as a mixed integer program whose objective function is convex.
Convex extensions that strike a practical balance between tightness and computational complexity are a topic of future work.

\bibliography{manuscript}

\end{document}